\documentclass[runningheads]{llncs}
\usepackage{booktabs}        
\usepackage{multirow}        
\usepackage{adjustbox}       
\usepackage{amsmath}         
\usepackage{booktabs}
\usepackage{multirow}
\usepackage{graphicx}
\usepackage{adjustbox}
\usepackage{longtable}
\usepackage{lscape} 
\usepackage{longtable}
\usepackage{graphicx}
\usepackage{adjustbox}
\usepackage{longtable}
\usepackage{array}
\usepackage{booktabs}
\usepackage{afterpage}
\usepackage[numbers,sort&compress]{natbib}
\usepackage{booktabs} 
\usepackage{caption}  
\usepackage{graphicx}    
\usepackage{multicol}    
\usepackage{float}       
\usepackage{subcaption}  
\usepackage{lipsum}      
\usepackage{makecell}
\usepackage{hhline}
\usepackage{hyperref}
\usepackage{pax}

\usepackage[T1]{fontenc}
\usepackage{graphicx}
\usepackage{xifthen}
\usepackage{hyperref}
\hypersetup{
colorlinks   = true,
citecolor    = blue,
urlcolor    =  blue
}
\usepackage{color}

\usepackage{tikz}
\usetikzlibrary{positioning,shapes.misc}
\usepackage{booktabs}
\usepackage{array}
\usepackage{hhline}
\usepackage{multirow}

\definecolor{tiPosCol}{rgb}{0.58, 0.77, 0.45}
\definecolor{tiNegCol}{rgb}{1.0, 0.41, 0.38}
\newcommand{\footURL}[1]{\footnote{\url{#1}}}

\newcommand{\stackedbar}[2]{%
    \begin{tikzpicture}[x=2.0cm, y=0.3cm] 
        \pgfmathsetmacro{\total}{#1+#2}
        \pgfmathtruncatemacro{\inttotal}{\total} 
        \ifnum\inttotal=0
            \draw[fill=gray!20] (0,0) rectangle (1,0.3);
        \else
            \pgfmathsetmacro{\fraction}{#1/\total}

            \pgfmathtruncatemacro{\intpartone}{#1} 
            \pgfmathtruncatemacro{\intparttwo}{#2} 
            \ifnum\intpartone>0
                \draw[fill=tiPosCol] (0,0) rectangle (\fraction,0.3);
                \pgfmathsetmacro{\xpos}{0.1} 
                \node[above] at (\xpos, 0.3) {\scriptsize \intpartone};
            \fi
            \ifnum\intparttwo>0
                 \draw[fill=tiNegCol] (\fraction,0) rectangle (1,0.3);
                \pgfmathsetmacro{\xpos}{0.9} 
                \node[above] at (\xpos, 0.3) {\scriptsize \intparttwo};
            \fi
        \fi
    \end{tikzpicture}%
}

\DeclareRobustCommand\TikzPos{\tikz \draw[fill=tiPosCol] (0,0) rectangle (0.5,0.8\baselineskip);}

\DeclareRobustCommand\TikzNeg{\tikz \draw[fill=tiNegCol] (0,0) rectangle (0.5,0.8\baselineskip);}

\usepackage{makecell}
\usepackage{dcolumn}

\begin{document}
\title{\texttt{GeeSanBhava}: Sentiment Tagged Sinhala Music Video Comment Data Set}
%
%
\author{Yomal De Mel\orcidID{0009-0005-3586-7899}\and
Nisansa de Silva\orcidID{0000-0002-5361-4810}}
\authorrunning{Y. De Mel and N. de Silva}
%
\institute{Department of Computer Science \& Engineering,\\ University of Moratuwa, Moratuwa, Sri Lanka\\
\email{\{mario.23,NisansaDdS\}@cse.mrt.ac.lk}\\
}
\maketitle              
\begin{abstract}
This study introduce \texttt{GeeSanBhava}, a high-quality data set of Sinhala song comments extracted from YouTube manually tagged using Russell's Valence-Arousal model by three independent human annotators. The human annotators achieve a substantial inter-annotator agreement (\textit{Fleiss' kappa} = 84.96\%).
The analysis revealed distinct emotional profiles for different songs, highlighting the importance of comment-based emotion mapping.  The study also addressed the challenges of comparing comment-based and song-based emotions, mitigating biases inherent in user-generated content. 
A number of Machine learning and deep learning models were pre-trained on a related large data set of Sinhala News comments in order to report the zero-shot result of our Sinhala YouTube comment data set. 
An optimized Multi-Layer Perceptron model, after extensive hyperparameter tuning, achieved a \textit{ROC-AUC} score of 0.887. The model is a three-layer MLP with a configuration of 256, 128, and 64 neurons. 
This research contributes a valuable annotated dataset and provides insights for future work in Sinhala Natural Language Processing and music emotion recognition. The complete data set is publicly available at: \url{https://bit.ly/SinhalaYoutubeComments}.

\end{abstract}
%
%
%
\section{Introduction}

Understanding and classifying emotions expressed on social media is crucial for various applications such as, sentiment analysis~\cite{jayawickrama2022facebook,jayakody2024aspect,jayakody2024enhanced,jayakody2024instruct}, product development~\cite{gunathilaka2025automatic}, social science research~\cite{haryanto2019facebook}, personalized recommendations~\cite{xu2022systematic}, and even mental health monitoring~\cite{stephen2019detecting,sarsam2021lexicon}. This paper presents a comprehensive study on emotion recognition in Sinhala song comments extracted from YouTube. Researchers leverage Russell's \textit{Valence-Arousal} (V-A) model~\cite{russell1980circumplex} to categorize emotions, providing a nuanced understanding beyond simple positive/negative sentiment.  This model allows us to capture the intensity (\textit{arousal}) and direction (\textit{valence}) of emotional responses, offering a richer representation of user experience with music.  By mapping comments to this two-dimensional space, it can visualize and analyze the emotional landscape associated with different songs.  This approach is particularly valuable in the context of music, where emotions are complex and multifaceted.

A key contribution of this work is the creation of, \texttt{GeeSanBhava}\footnote{The name \texttt{GeeSanBhava} is a concatenation of three transliterated Sinhala words: \textit{Gee} means Songs, \textit{San} is the shortened form of \textit{Sanna} meaning comment, and \textit{Bhava} means sentiment. Our complete data set is publicly available at: \url{https://bit.ly/SinhalaYoutubeComments}
}, a human annotated dataset of 63,471 Sinhala song comments, labeled across a spectrum of emotions. A rigorous annotation process was followed, involving three independent annotators who achieved a high inter-annotator agreement (\textit{Fleiss's kappa} = $84.96\%$), which ensures the reliability of our data set.  
This data set is, to the best of our knowledge, the largest and most comprehensively annotated data sets for Sinhala sentiment and emotion analysis using Russell's Valence-Arousal mapping. 
In addition, the relationship between emotions expressed in user comments and the perceived emotion of the songs themselves were explored. Comparison of emotion categorizations based on comments and songs using \textit{cosine similarity}. 
User comments, while providing valuable information on listener reactions, can be influenced by various factors, including individual preferences, social context, and even the presence of other comments.  By comparing these emotional responses with expert annotations of the songs' emotional content, it was aimed to understand the interplay between musical features and subjective listener perceptions.  This comparison helped identify potential discrepancies and biases, leading to a more nuanced understanding of how music evokes emotions.  
Finally, the study developed and evaluated several machine learning and deep learning models for classifying Sinhala comments, achieving the most promising results with a Multi-Layer Perceptron (MLP) 
which yielded a \textit{ROC-AUC} score of 0.887. 

\section{Related Work}

Music Emotion Recognition (MER) is a subfield of Music Information Retrieval (MIR) that applies computational techniques to analyze and classify the emotional content of music~\cite{hizlisoy2021music}. It has broad applications, including music recommendation, mood-based playlist generation, and affective computing~\cite{kim2010music}. However, this field faces multiple challenges such as the subjective nature of emotions and the lack of a standardized classification, which complicates the development of accurate models~\cite{yang2012machine}. The emotional impact of music is influenced by structural features that can be identified as melody, harmony, rhythm, and timbre, which interact with human cognitive and affective processes~\cite{thoma2012emotion}. While interdisciplinary research has led to various emotion models and classifications, challenges remain in gaining reliable annotations and capturing the temporal aspects of music emotion~\cite{can2023approaches}. Addressing these issues requires expertise from signal processing, machine learning, psychology, and musicology. Recent advancements in deep learning have enhanced Natural Language Processing (NLP) through word embeddings, which map words to dense vector representations~\cite{lai2015recurrent}. Word2Vec, introduced by~\citet{mikolov2013efficient}, has been foundational,
to capture semantic relationships. These techniques, despite ignoring word order, have significantly contributed to the improvement in various NLP tasks, including sentiment and emotion analysis~\cite{mikolov2013distributed}.

Sinhala, an Indo-Europian language spoken as L1 only by a small ethnic group largely limited to the island of Sri Lanka, has only very few NLP resources created for it and is thus considered a Low-Resourced Language~\cite{de2019survey}. 
\citet{ranathunga2021sentiment} make the claim that this has especially hindered the progress in areas such as sentiment analysis for Sinhala. To alleviate this, they introduce 
a publicly accessible corpus for sentiment polarity annotation in Sinhala language using News comments.  Each News comment within the corpus is labeled as \texttt{POSITIVE}, \texttt{NEGATIVE}, or \texttt{NEUTRAL}, enabling sentiment classification at the document level.

Russell’s Valence-Arousal (V-A) model~\cite{russell1980circumplex} is a well-established framework in affective computing and emotion analysis. This model conceptualizes emotions within a two-dimensional space, where valence represents a range from positive to negative sentiment, and arousal ranges from calm to highly energetic emotional states.

\section{Methodology}

\subsection{Dataset and Emotion Mapping of Sinhala Song Comments}
\label{sec:meth:com}

The base dataset utilized in this study contains Sinhala-language comments extracted from YouTube videos of Sinhala songs by~\citet{de2025linguistic}. 
These comments were systematically subjected to an independent annotation process conducted by three independent annotators to minimize subjective biases and establish consensus-driven categorizations. The annotation process was guided by the Russell’s Valence-Arousal (V-A) model~\cite{russell1980circumplex}.
The inclusion of multiple perspectives in the annotation process enhances the reliability and validity of the emotion labels, making this dataset a valuable resource for computational models in sentiment analysis and emotion recognition. The annotated emotions cover a wide range of feelings often expressed in text, from high-energy positive emotions to low-energy negative emotions. However since the research focuses on the Sinhala YouTube comments on Sinhala song videos this would limit the generalizability to other languages. Sinhala comments  reflect culturally specific ways of expressing emotion and sentiment and these are with unique linguistic features that differ significantly from more other languages. Further more additional information about the dataset source diversity (e.g., genres, artists) and comment characteristics (e.g., average length, language distribution) are available in the dataset repository on GitHub and on the study done by~\citet{de2025linguistic}.

\subsection{Comparison Against the Emotion Mapping of Sinhala Songs}
\label{sec:meth:son}

As the next step in the research, three independent annotators were tasked to categorise the emotions of each song found in the data set (as opposed to the comments as in Section~\ref{sec:meth:com}) using Russell’s emotion model. The annotators were provided with the audio recordings of the songs and their corresponding lyrics to ensure a comprehensive understanding of the emotional essence conveyed by both the melody and lyrical content. This manual annotation process allowed for an independent assessment of the emotional category of each song. First, a simple direct comparison was conducted between the highest frequency emotion vector derived from the comments with the dominant emotion assigned to the song itself by the annotators using cosine
Similarity.

However, this direct comparison only considers the highest frequently occurring emotion for each song's comments, which does not consider the full spectrum of emotional diversity expressed by listeners. In order to address this limitation,  weighted average emotion vector for each song were computed, taking into account the relative frequency of each emotion category assigned in the comment dataset, calculating the centroid emotion vector for each annotator, and the centroid emotion vector for each song based on the comment dataset. This approach allowed for a more holistic representation of the emotional distribution within the comment-based data. The centroid calculation provided a more balanced representation of the overall emotional profile by aggregating and averaging the emotional responses rather than depending only on the most frequently occurring category.

Despite these refinements, inconsistencies are likely to raise between the comment-based vectors and the song-based vectors, due to the presence of inherent biases in user-generated content. To mitigate these biases and ensure a fair comparison, the comment-based emotion vectors were standardized using Equation~\ref{E1}. The standardization process involved normalizing the emotion vectors to a uniform scale, reducing the potential influence of extreme or outlier values.

\begin{equation}
\label{E1}
c_{z,i} = \frac{c_i - \frac{\sum_{j=0}^{n} c_j}{n}}{\max_{k=0 \to n} \left(c_k - \frac{\sum_{j=0}^{n} c_j}{n}\right)}
\end{equation}

\subsection{Feature Engineering and Model Training for Sinhala Comment Classification}

To develop a classification model for Sinhala YouTube comments on Sinhala songs, the initial approach involved fine-tuning an existing model on a large data set. Given that our Sinhala YouTube comment dataset is new, the idea is to use the established related data set to pre-train a Sinhala comment classification model and obtain the zero-shot results on our YouTube comment dataset.
Upon reviewing related studies, the research conducted by~\citet{ranathunga2021sentiment} was identified as the closest reference, as it examined Sinhala news comments and classified them into \textit{negative}, \textit{positive}, and \textit{neutral} categories. However, the publicly available dataset associated with this study\footURL{https://github.com/theisuru/sentiment-tagger/tree/master/corpus} contained only the comments classified as \textit{negative} and \textit{positive}. Further, the trained models from their work were also not accessible. As a result, the study attempted to replicate their work as the baseline by following the methodology they have outlined while making necessary modifications to improve adaptability for social media comments.

The first step in this process was data preprocessing, which involved refining the dataset to ensure consistency and remove unnecessary elements. Non-Sinhala characters such as numbers, symbols, and special characters were eliminated to standardize the text format. Additionally, missing or incomplete data were handled appropriately to prevent inconsistencies in the training process. 

The next phase of the study involved converting the preprocessed comments into numerical representations using different embedding techniques. Four pre-trained embedding models were used: word2vec~\cite{mikolov2013efficient}, fastText~\cite{bojanowski2017enriching}, SinBERT~\cite{dhananjaya2022bertifying}, and sentence embeddings (SBERT)~\cite{reimers-2019-sentence-bert}. These models captured the semantic meaning of words and phrases, enabling more effective classification. As suggested in the baseline study~\cite{ranathunga2021sentiment}, the mean vector of the word embeddings was used to generate a single vector representation for each comment. In addition to embeddings, TF-IDF features were extracted from the textual data to capture the significance of words within the comments. 

Once the features were extracted, multiple traditional machine learning classifiers were employed to train and evaluate the models. These included Logistic Regression, Random Forest, Support Vector Machine (SVM), and XGBoost. The performance of these classifiers was tested using various combinations of embedding techniques and TF-IDF features. To ensure fair comparison and prevent bias, the dataset was divided into training and test sets using stratified sampling, maintaining the proportional distribution of labels. 
(Hyperparameter details are given in Appendix~\ref{app:Hype})
Given the high dimensionality of embedding-based feature representations, dimensionality reduction was applied using Truncated Singular Value Decomposition (SVD). 
Each classifier was trained on different feature sets, and performance metrics such as \textit{accuracy}, \textit{precision}, \textit{recall}, and \textit{F1-score} were used to evaluate their effectiveness. 

\section{Results and Discussion}

\subsection{Emotion Mapping of Sinhala Song Comments}

\begin{figure}[!htb]
\centering
\resizebox{\textwidth}{!}{
\begin{tabular}{ccc|ccc}
    \textbf{Annotator 1} & \textbf{Annotator 2} & \textbf{Annotator 3}& \textbf{Annotator 1} & \textbf{Annotator 2} & \textbf{Annotator 3} \\ 
    \midrule
    \includegraphics[width=0.18\textwidth]{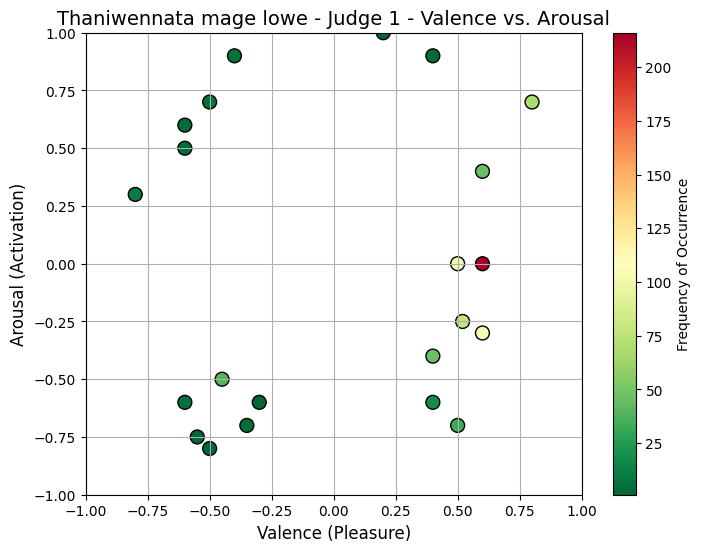} & 
    \includegraphics[width=0.18\textwidth]{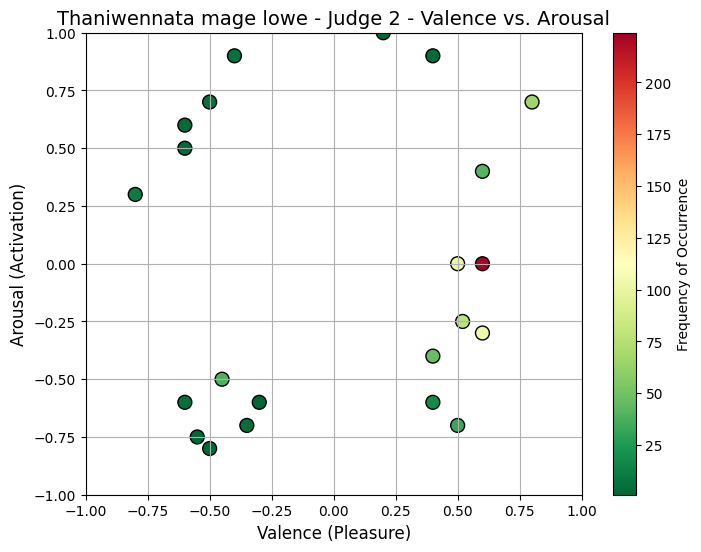} & 
    \includegraphics[width=0.18\textwidth]{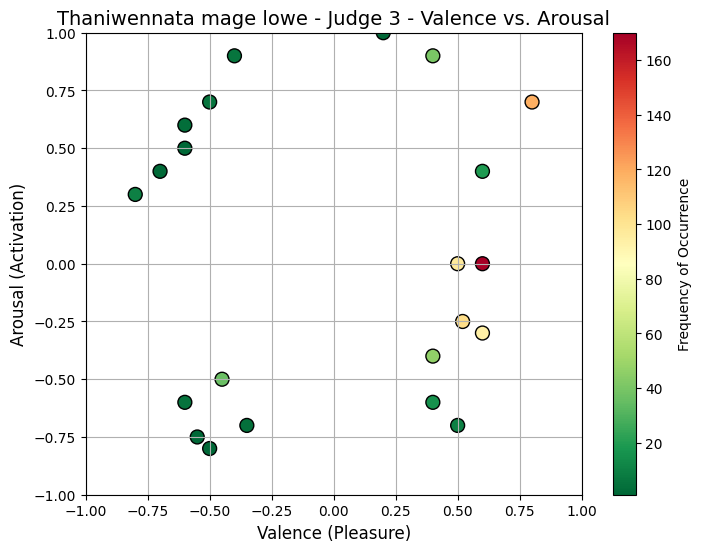} & 
    \includegraphics[width=0.18\textwidth]{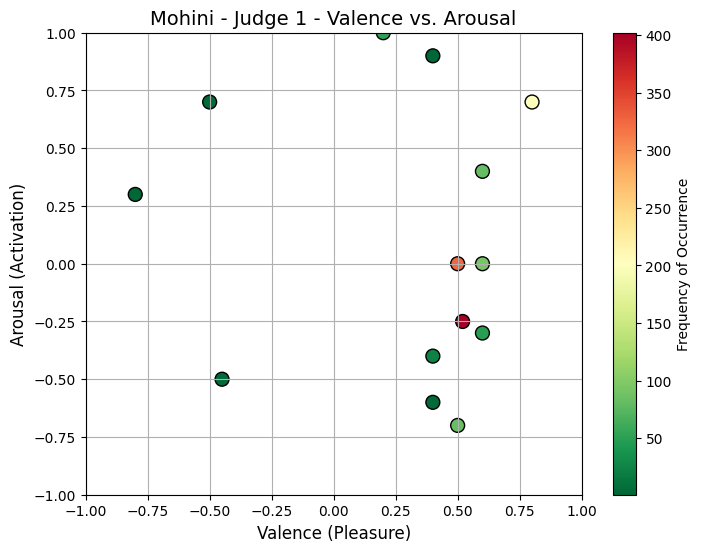} & 
    \includegraphics[width=0.18\textwidth]{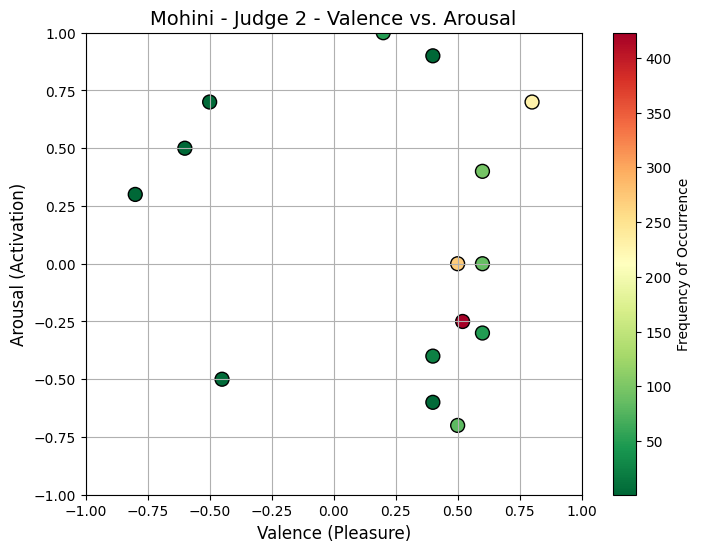} & 
    \includegraphics[width=0.18\textwidth]{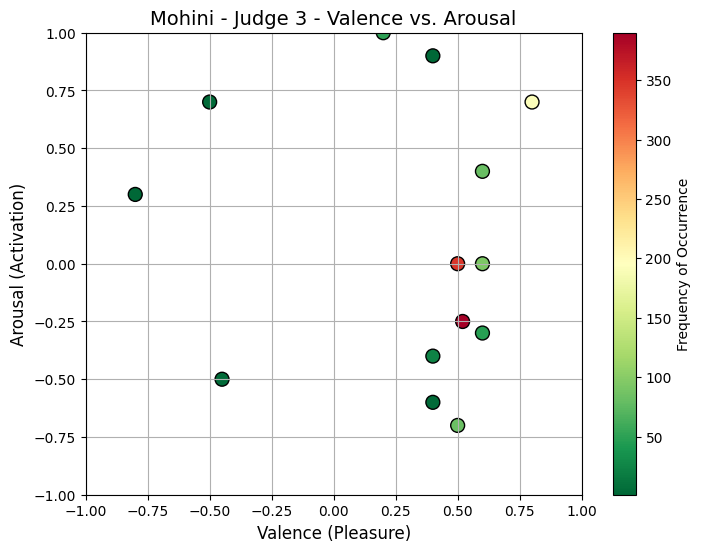} \\
    
    \includegraphics[width=0.18\textwidth]{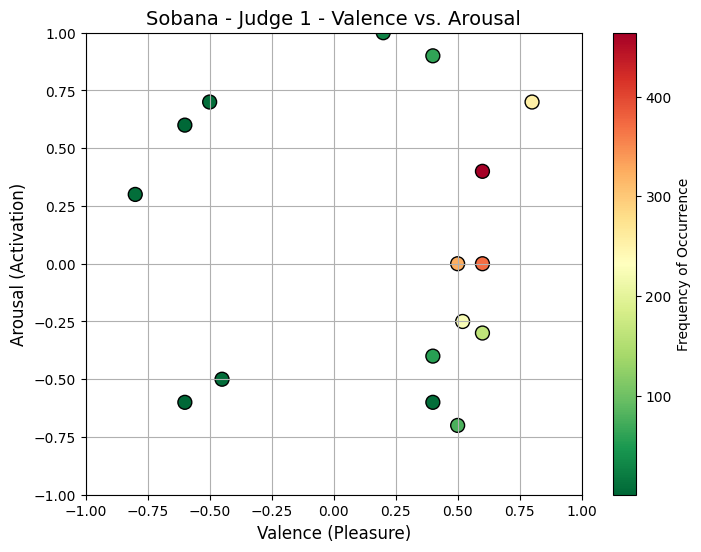} & 
    \includegraphics[width=0.18\textwidth]{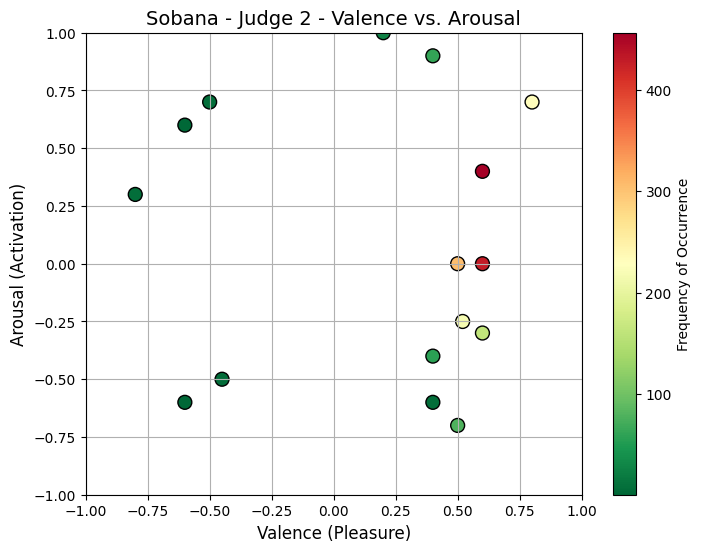} & 
    \includegraphics[width=0.18\textwidth]{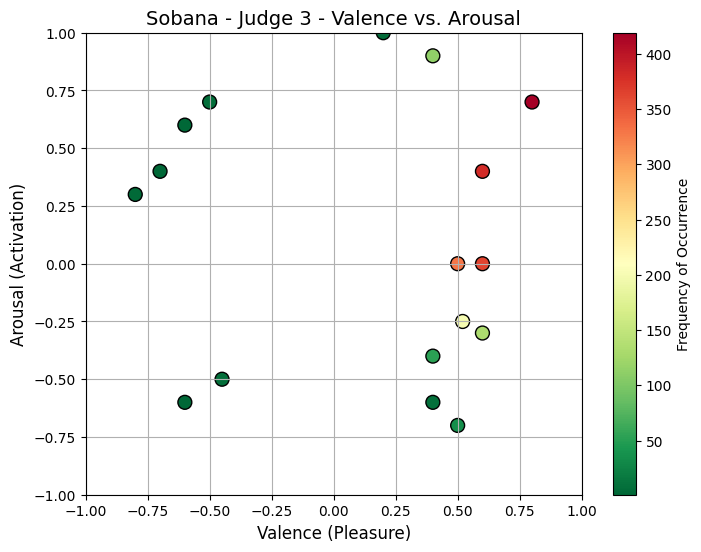} &
    \includegraphics[width=0.18\textwidth]{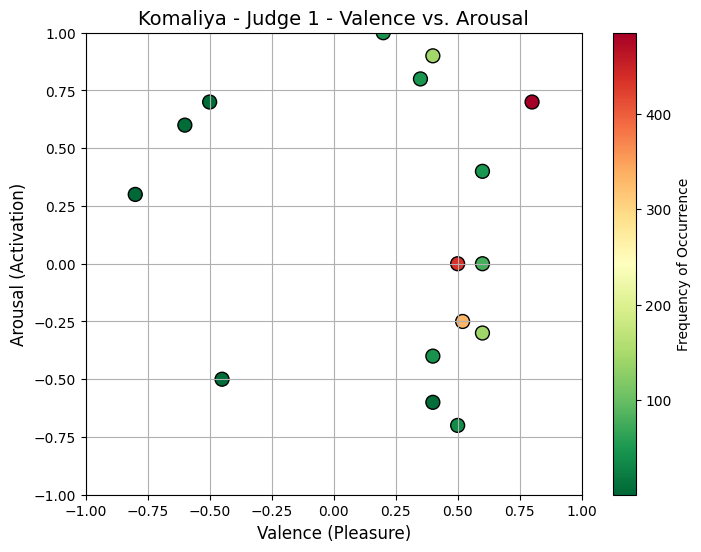} & 
    \includegraphics[width=0.18\textwidth]{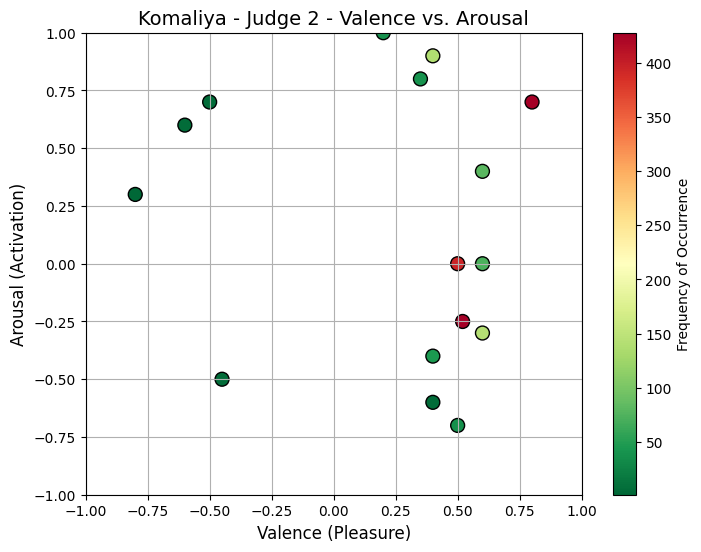} & 
    \includegraphics[width=0.18\textwidth]{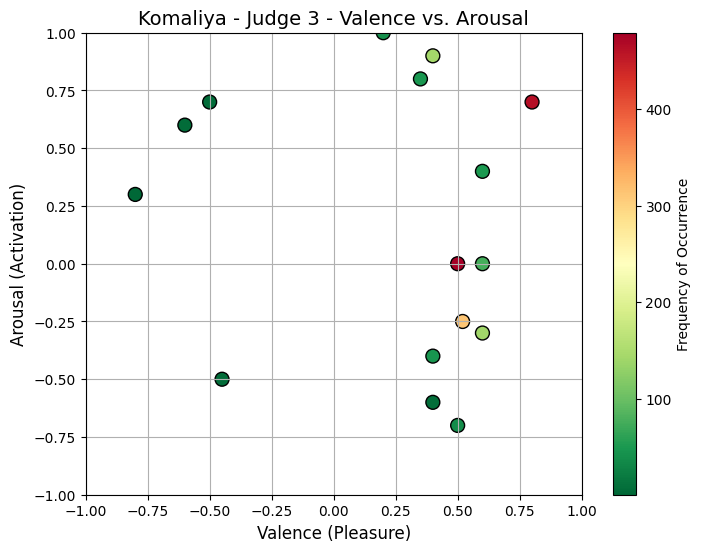} \\

    \includegraphics[width=0.18\textwidth]{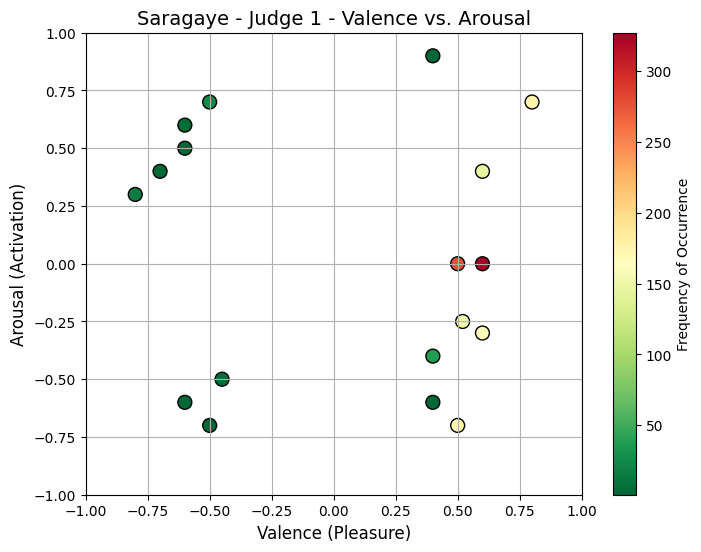} & 
    \includegraphics[width=0.18\textwidth]{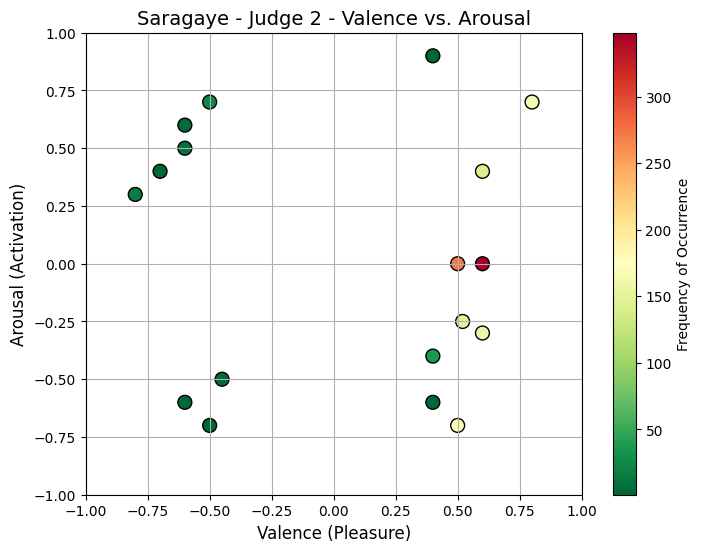} & 
    \includegraphics[width=0.18\textwidth]{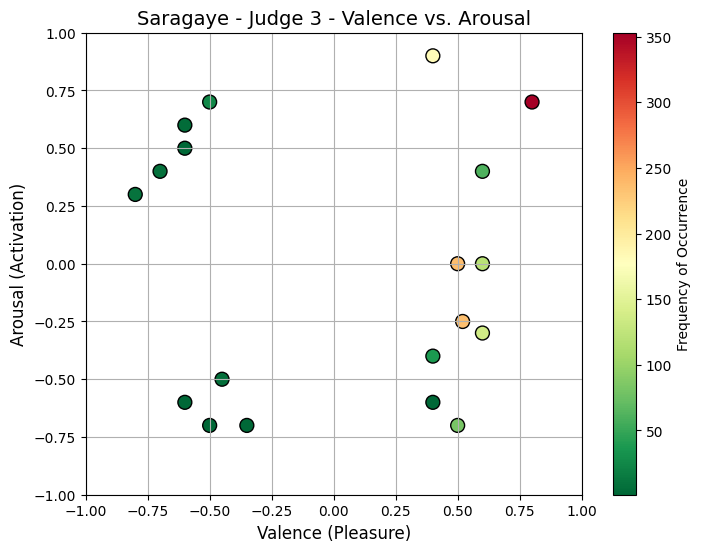} &
    \includegraphics[width=0.18\textwidth]{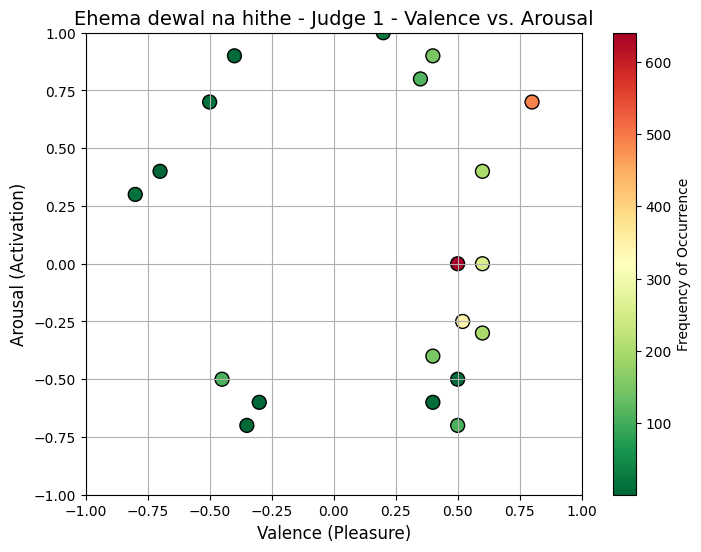} & 
    \includegraphics[width=0.18\textwidth]{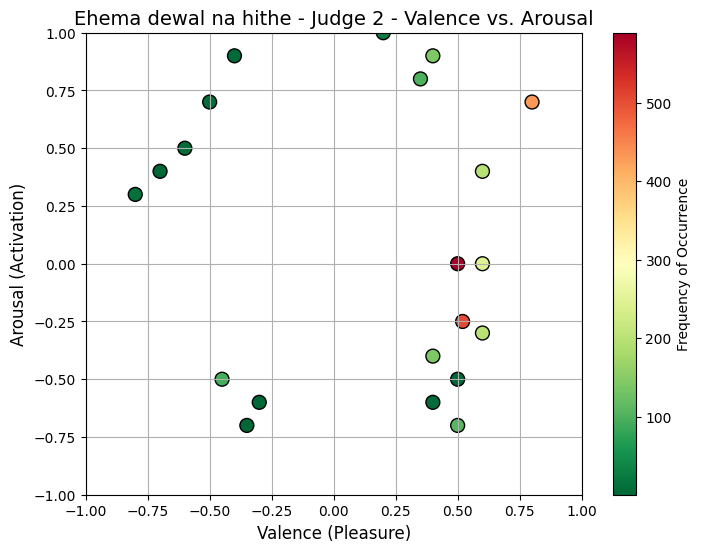} & 
    \includegraphics[width=0.18\textwidth]{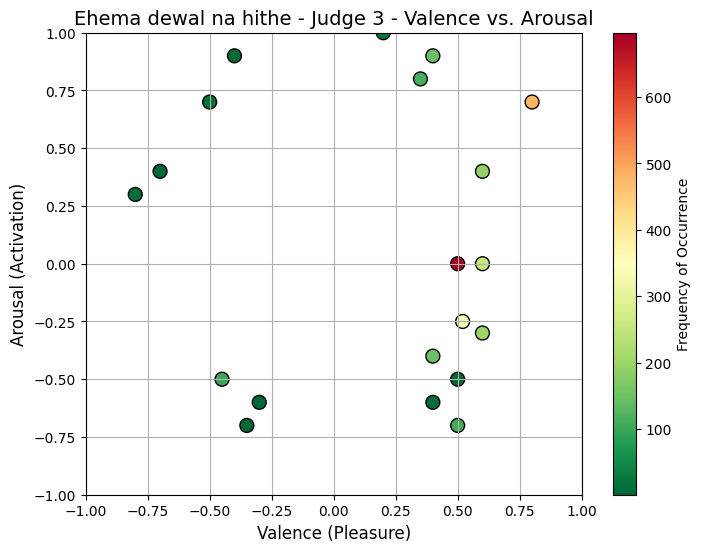} \\

    \includegraphics[width=0.18\textwidth]{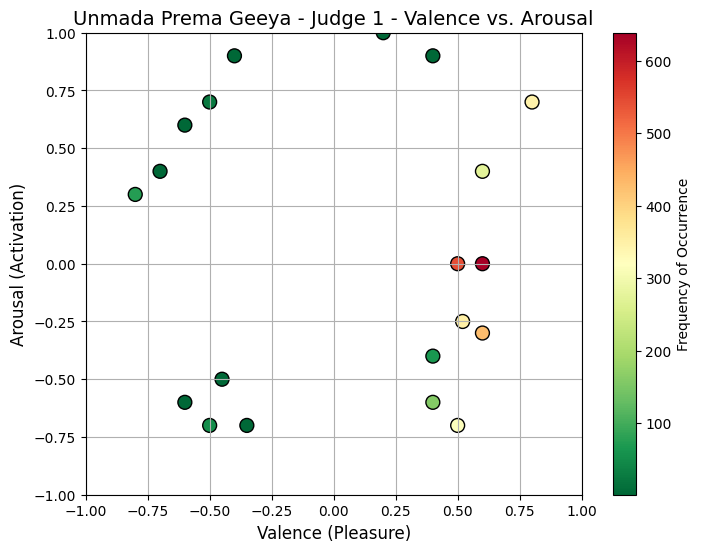} & 
    \includegraphics[width=0.18\textwidth]{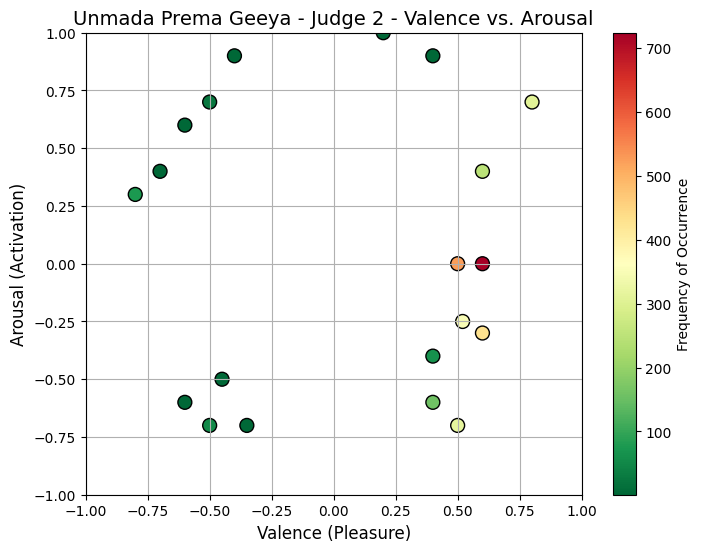} & 
    \includegraphics[width=0.18\textwidth]{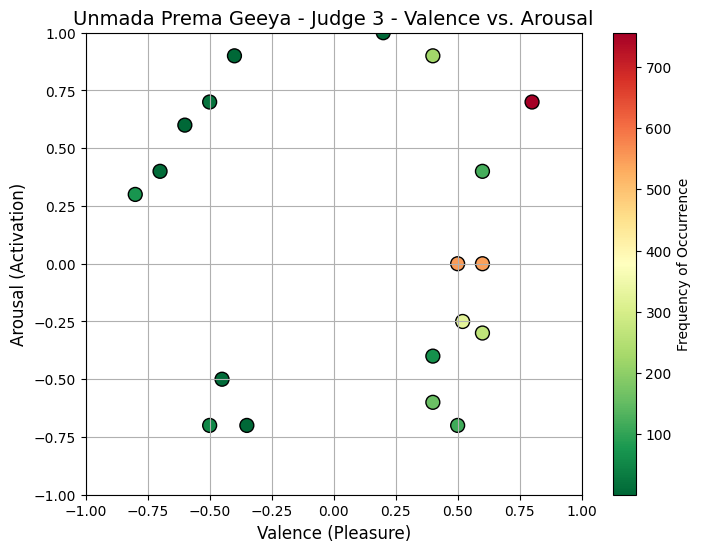} &

    \includegraphics[width=0.18\textwidth]{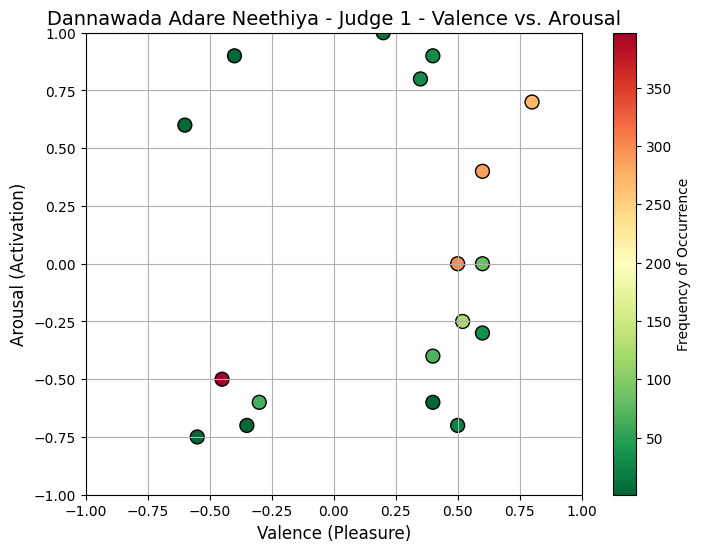} & 
    \includegraphics[width=0.18\textwidth]{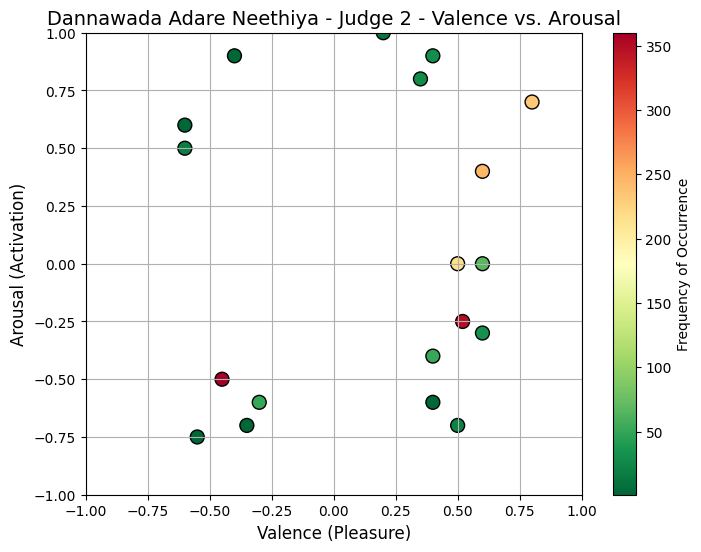} & 
    \includegraphics[width=0.18\textwidth]{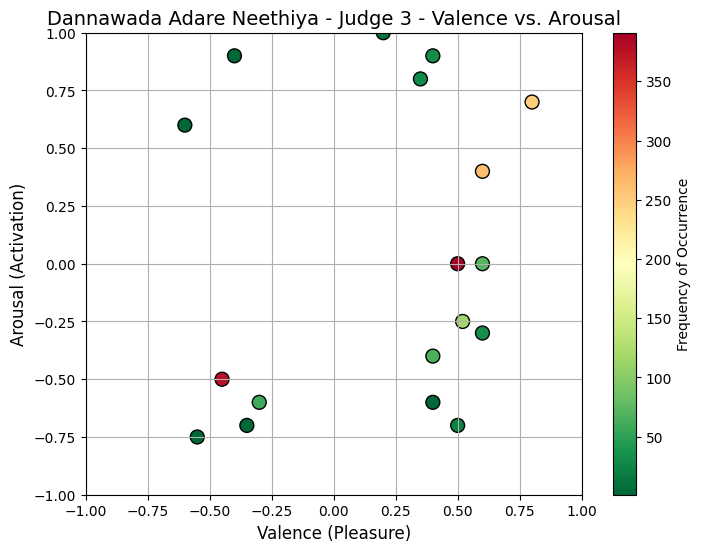} \\

    \includegraphics[width=0.18\textwidth]{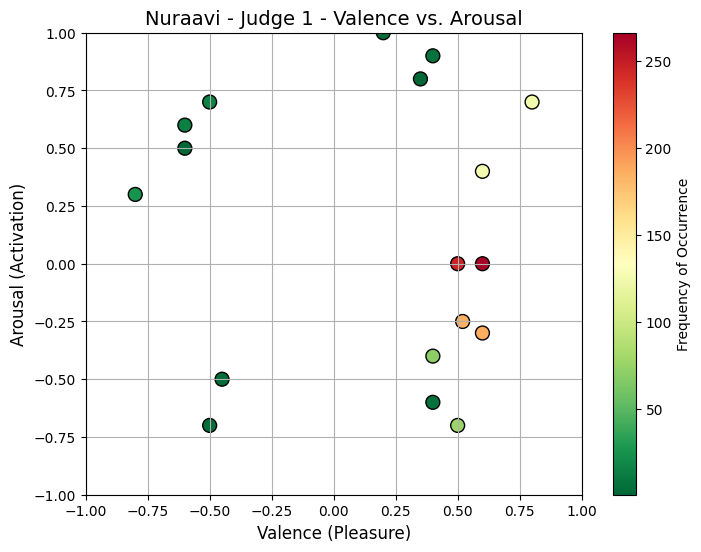} & 
    \includegraphics[width=0.18\textwidth]{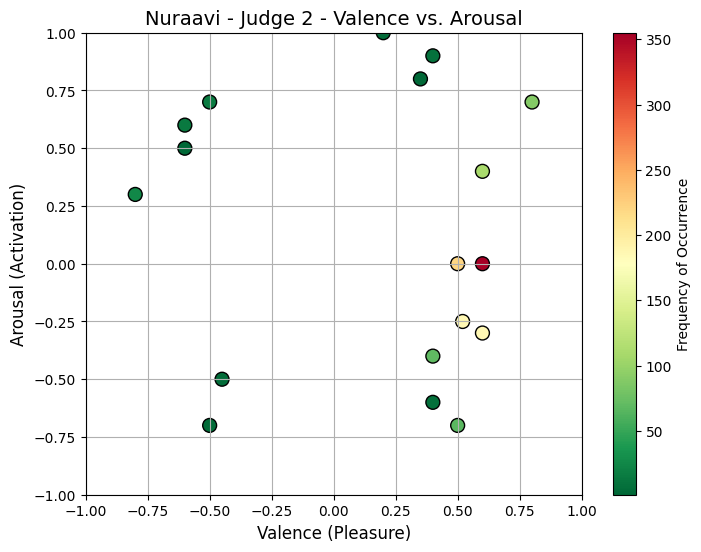} & 
    \includegraphics[width=0.18\textwidth]{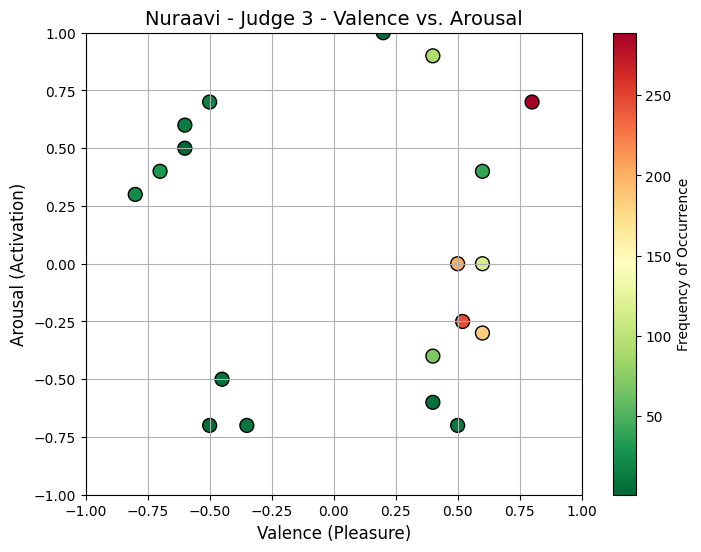} & 
    \includegraphics[width=0.18\textwidth]{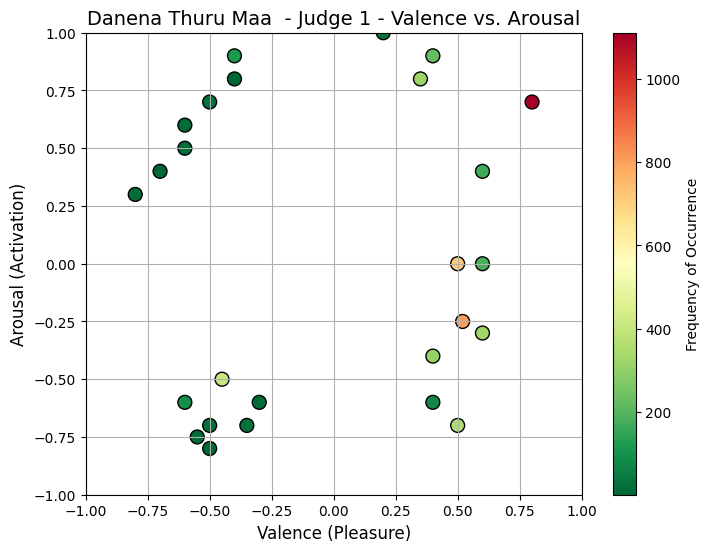} & 
    \includegraphics[width=0.18\textwidth]{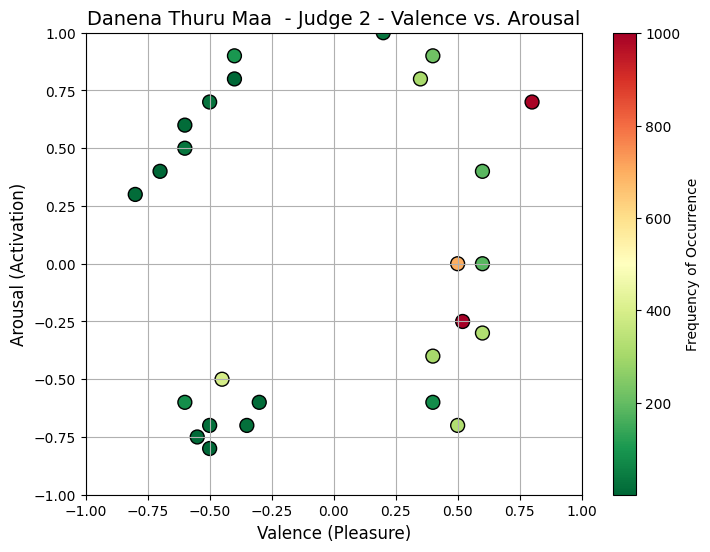} & 
    \includegraphics[width=0.18\textwidth]{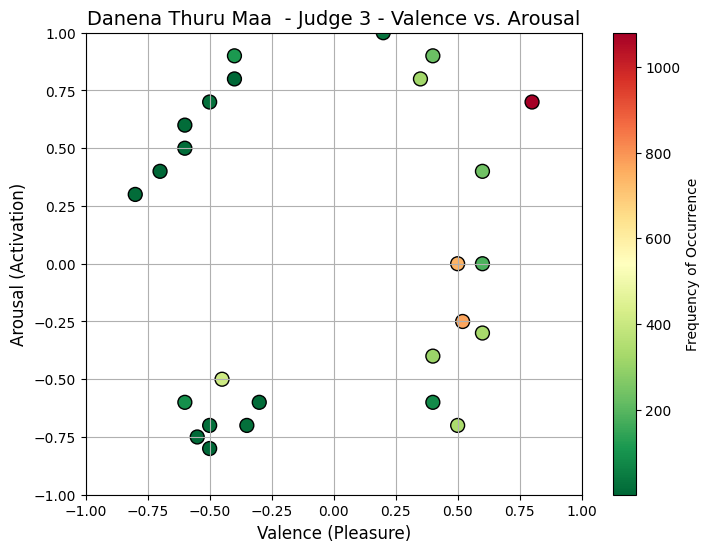} \\

    \includegraphics[width=0.18\textwidth]{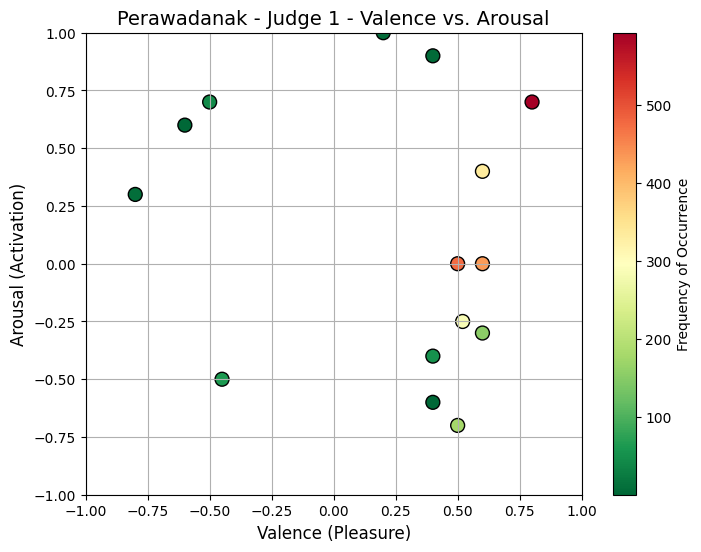} & 
    \includegraphics[width=0.18\textwidth]{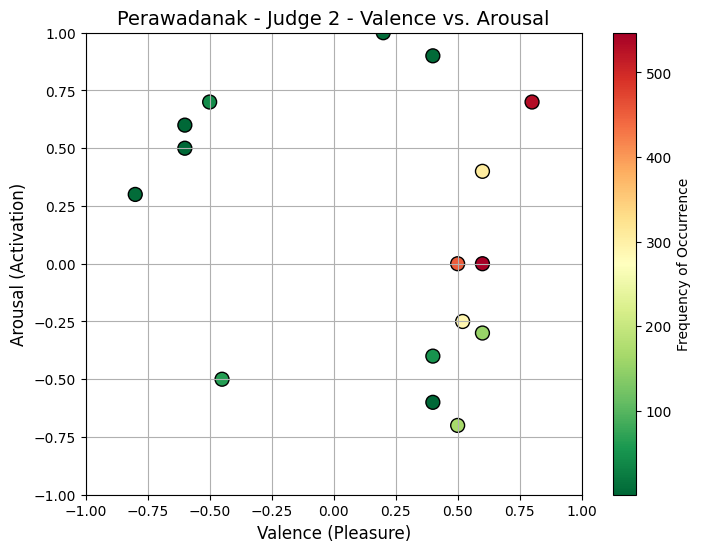} & 
    \includegraphics[width=0.18\textwidth]{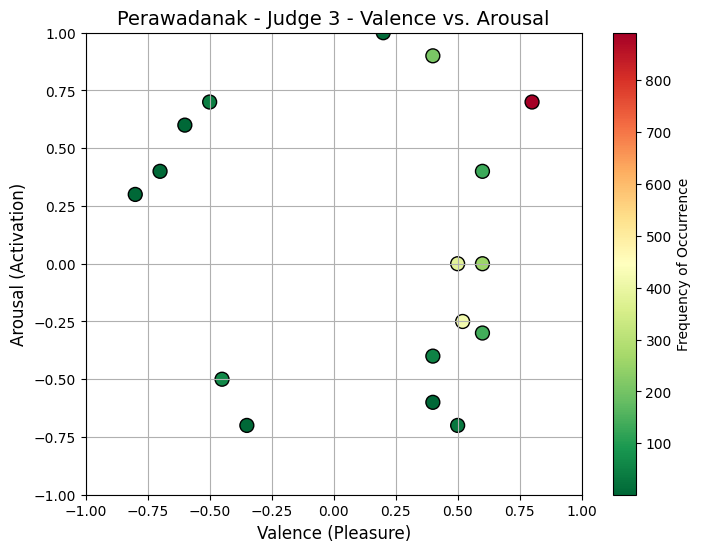} &

    \includegraphics[width=0.18\textwidth]{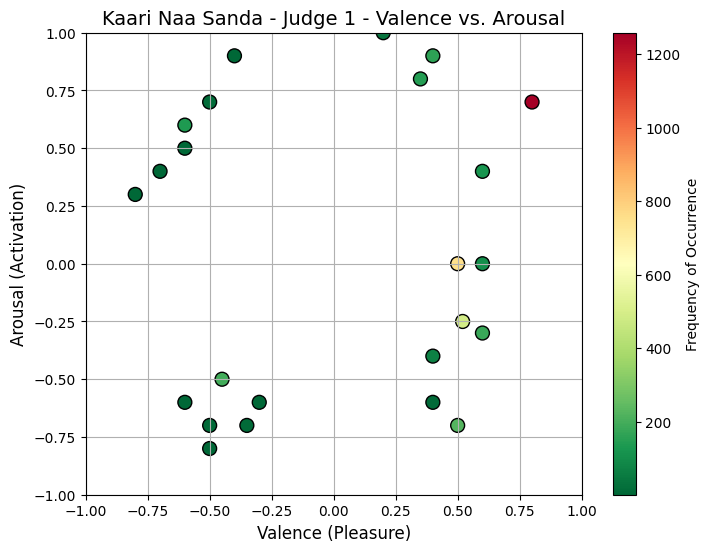} & 
    \includegraphics[width=0.18\textwidth]{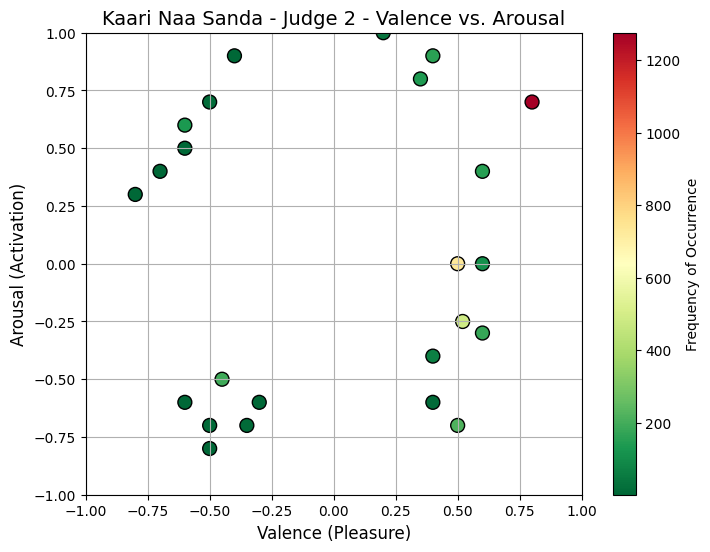} & 
    \includegraphics[width=0.18\textwidth]{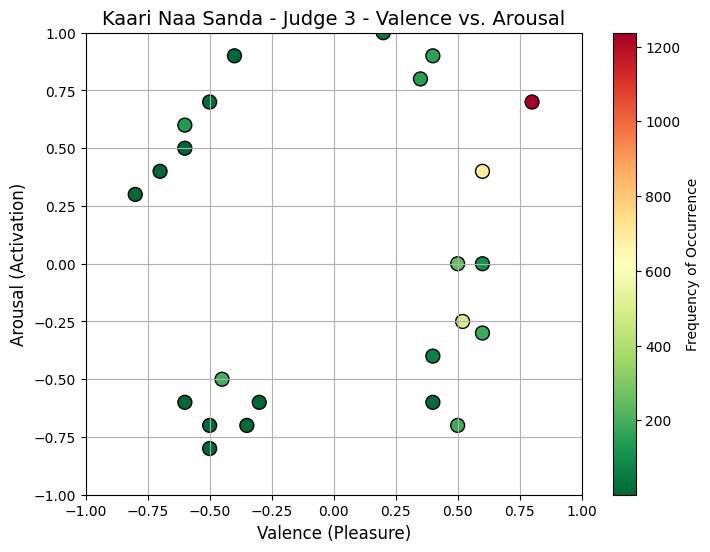} \\

    \includegraphics[width=0.18\textwidth]{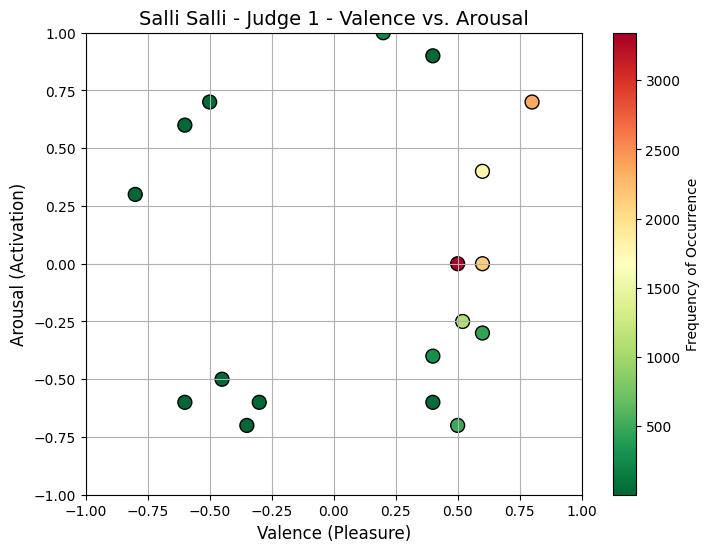} & 
    \includegraphics[width=0.18\textwidth]{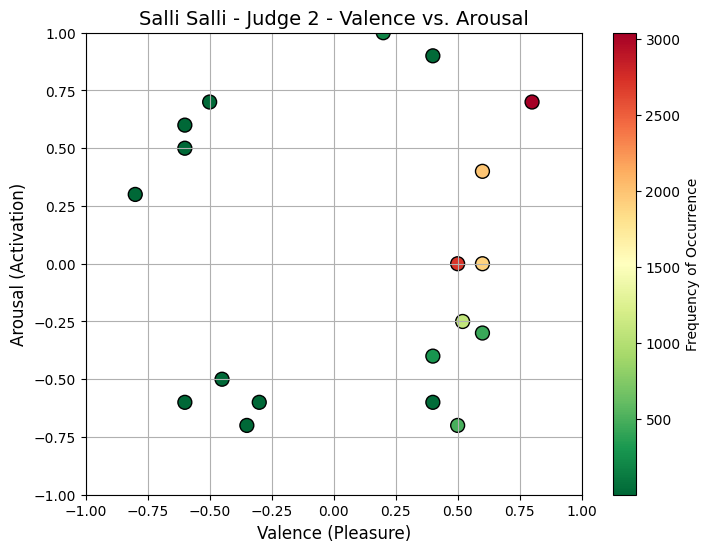} & 
    \includegraphics[width=0.18\textwidth]{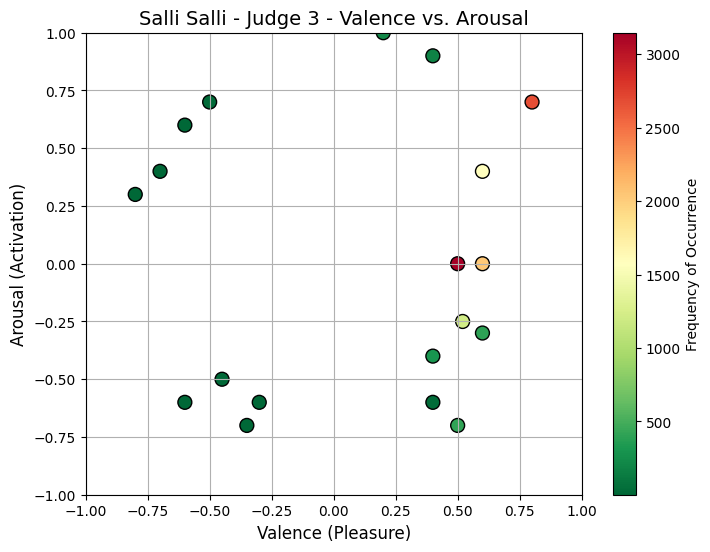} &

    \includegraphics[width=0.18\textwidth]{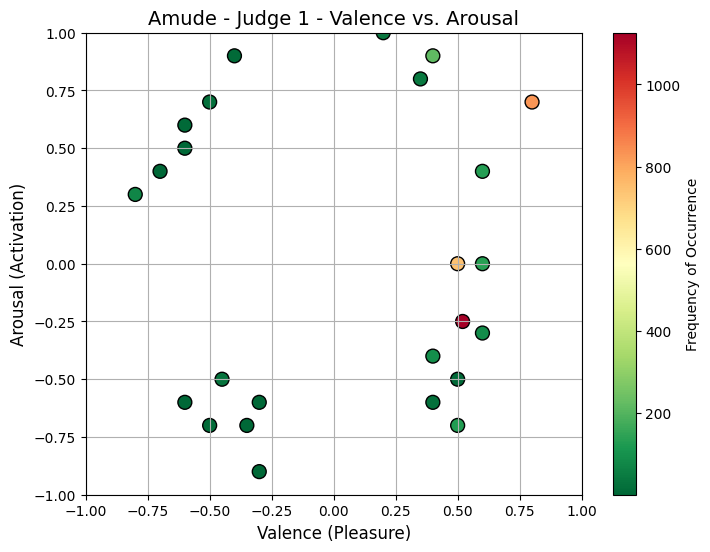} & 
    \includegraphics[width=0.18\textwidth]{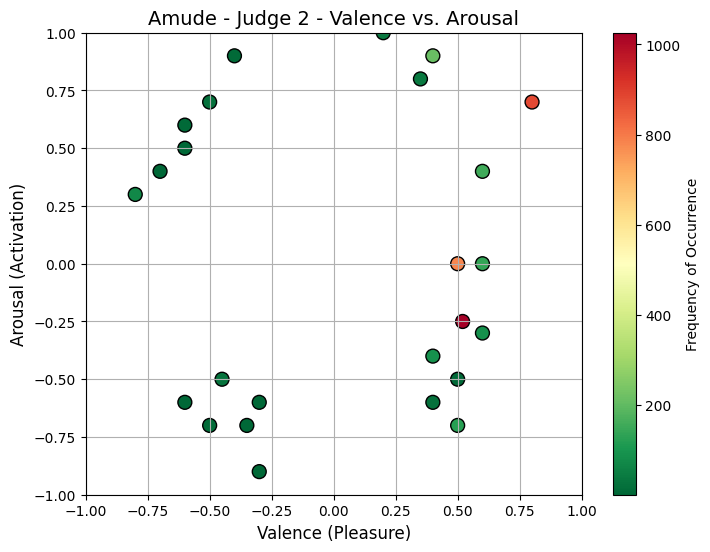} & 
    \includegraphics[width=0.18\textwidth]{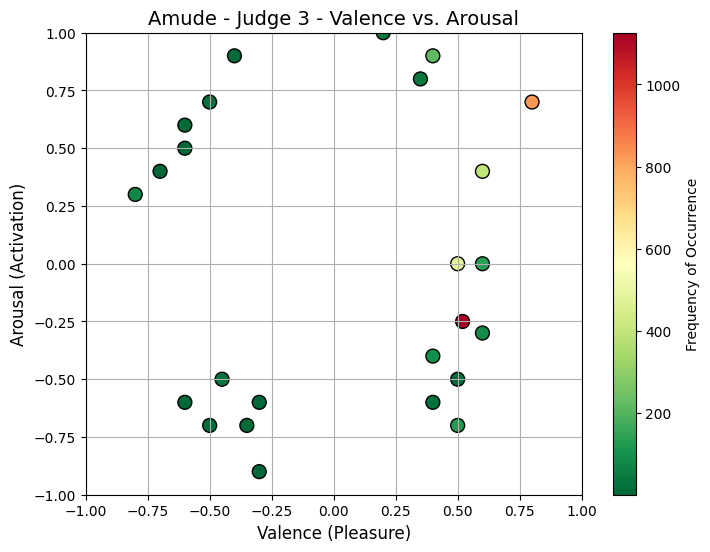} \\

    \includegraphics[width=0.18\textwidth]{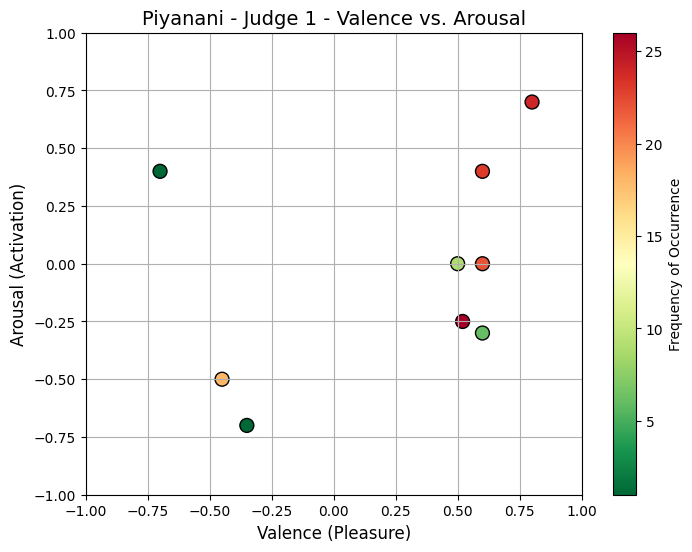} & 
    \includegraphics[width=0.18\textwidth]{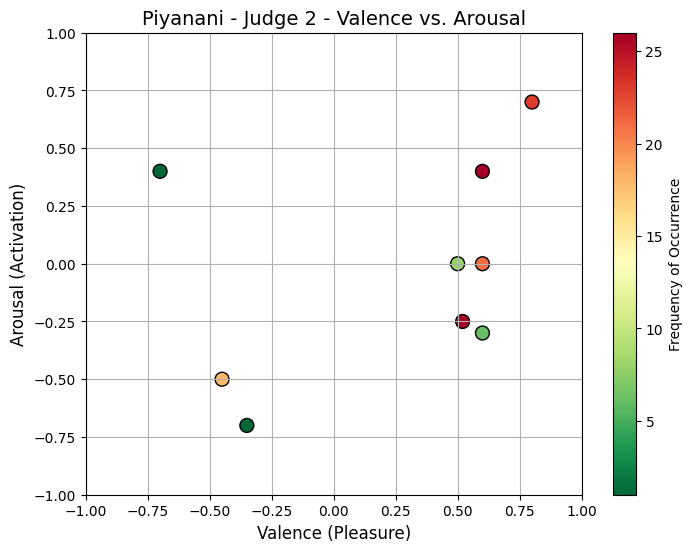} & 
    \includegraphics[width=0.18\textwidth]{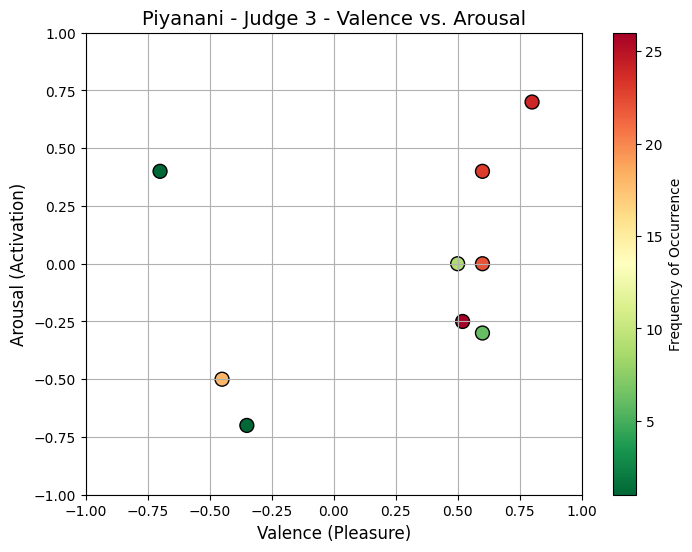} &
    \includegraphics[width=0.18\textwidth]{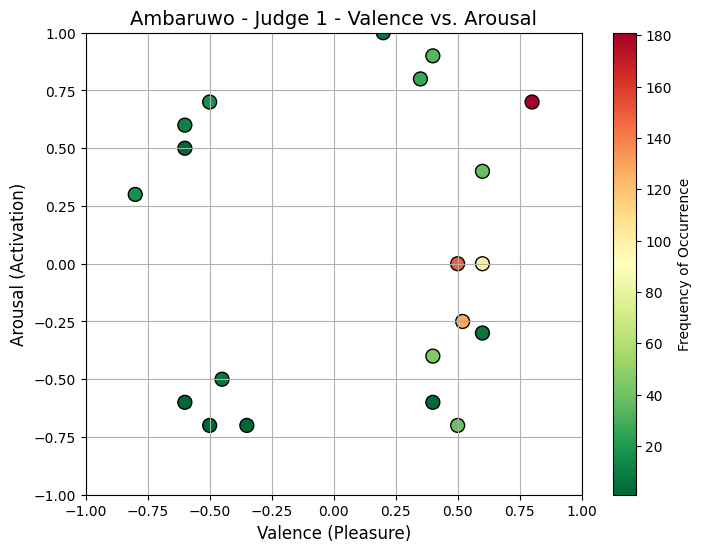} & 
    \includegraphics[width=0.18\textwidth]{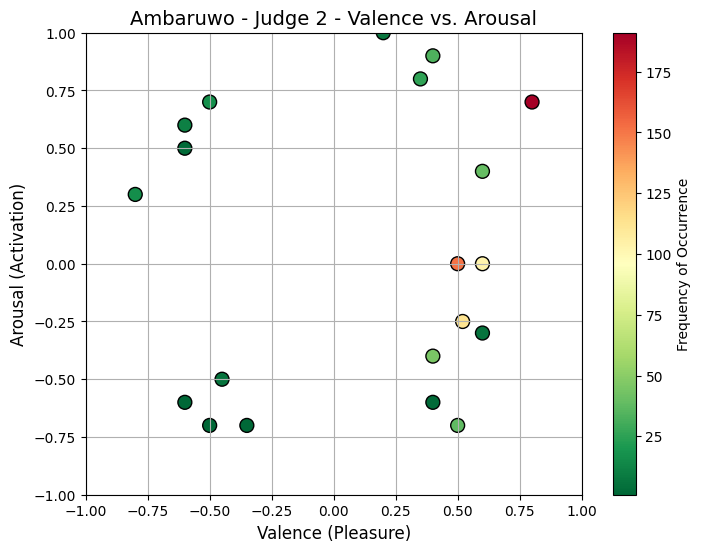} & 
    \includegraphics[width=0.18\textwidth]{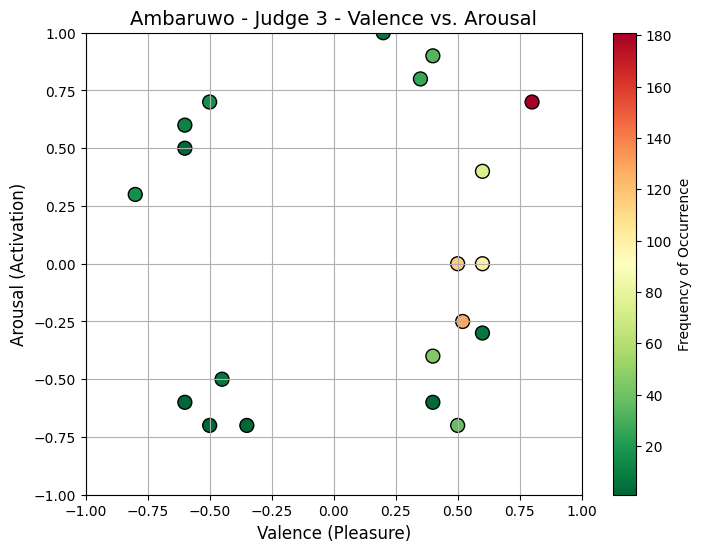} \\

    \includegraphics[width=0.18\textwidth]{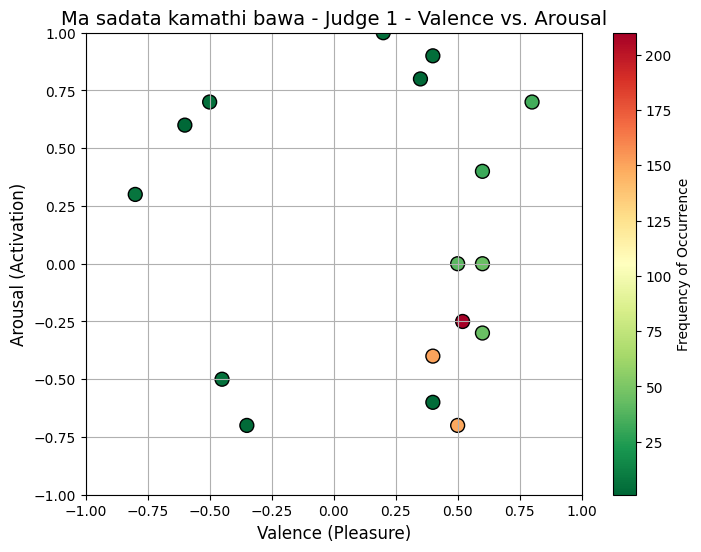} & 
    \includegraphics[width=0.18\textwidth]{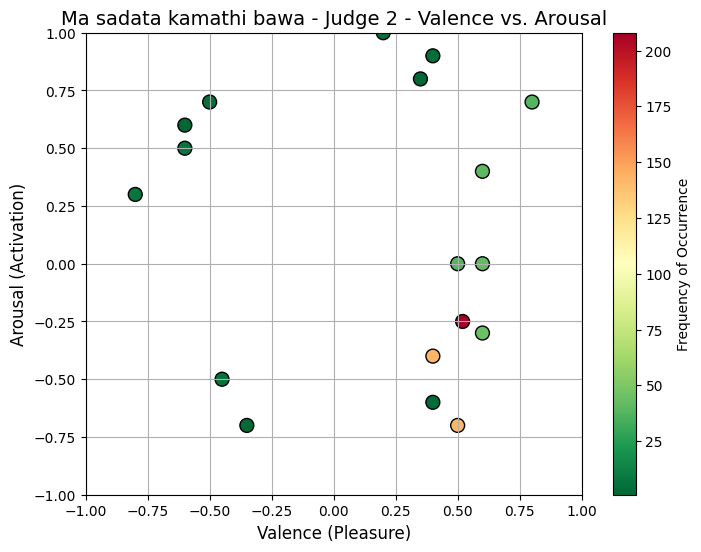} & 
    \includegraphics[width=0.18\textwidth]{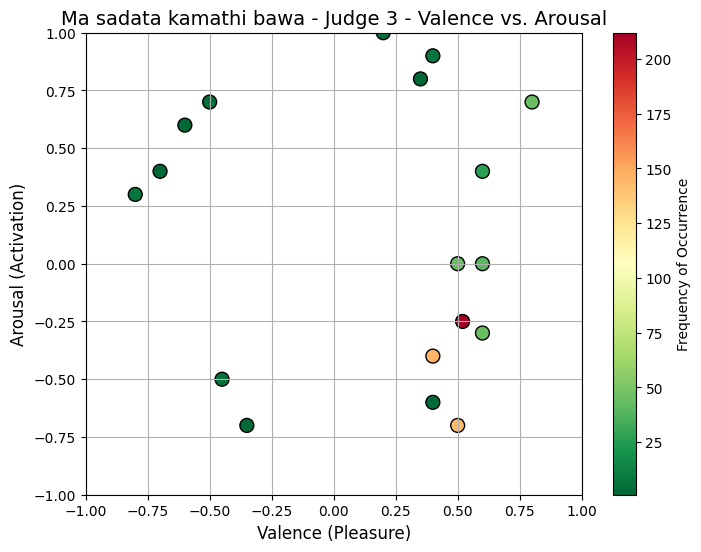} &

    \includegraphics[width=0.18\textwidth]{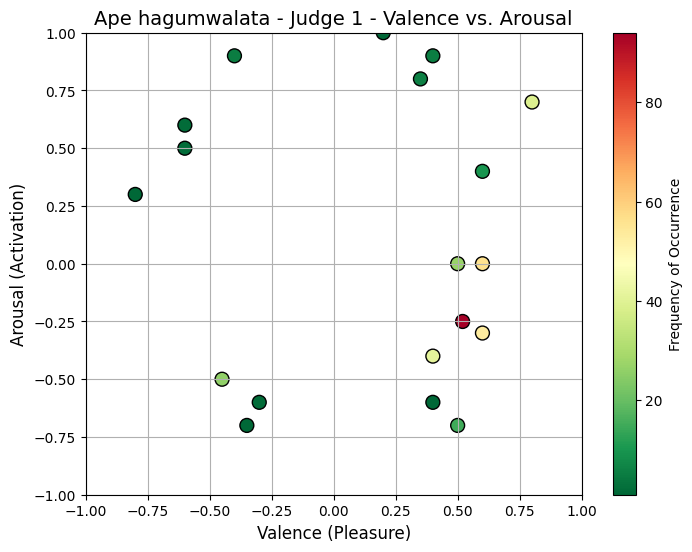} & 
    \includegraphics[width=0.18\textwidth]{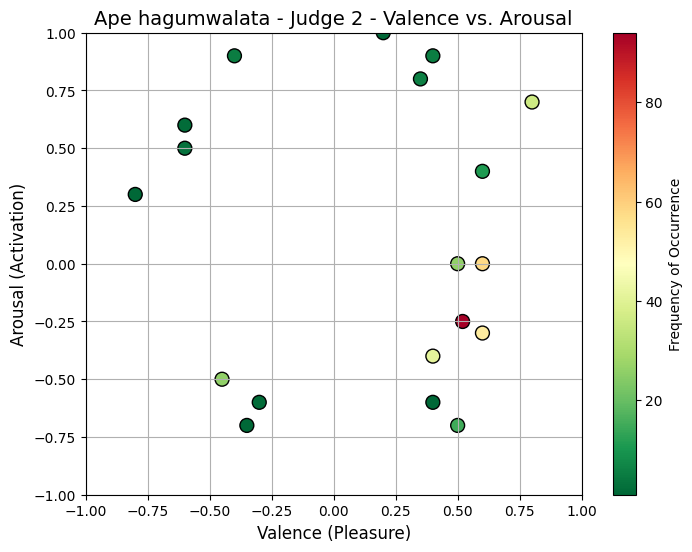} & 
    \includegraphics[width=0.18\textwidth]{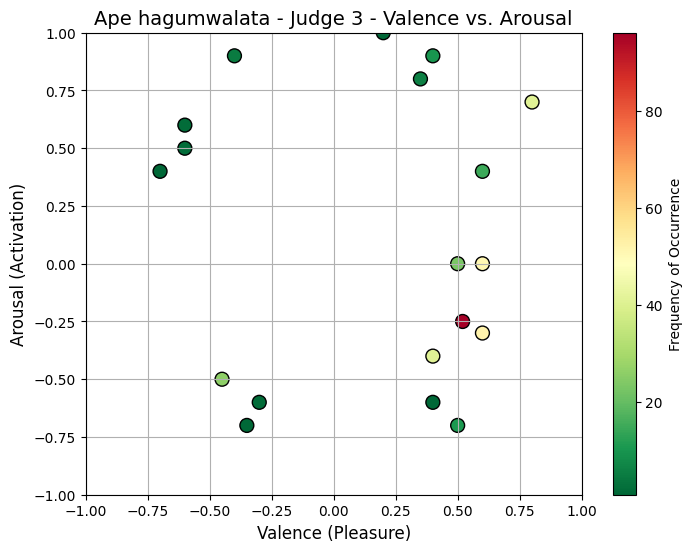} \\

    \includegraphics[width=0.18\textwidth]{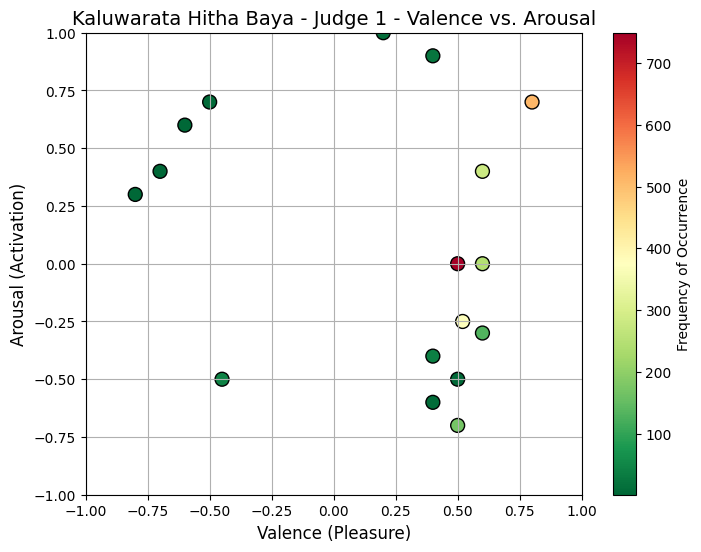} & 
    \includegraphics[width=0.18\textwidth]{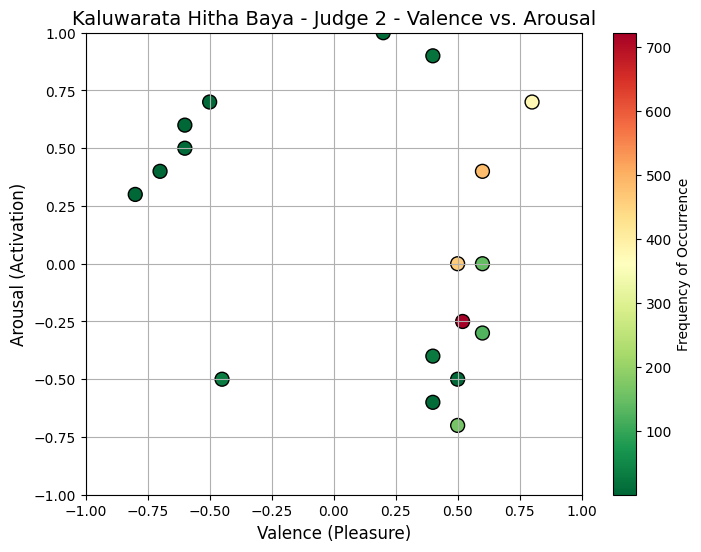} & 
    \includegraphics[width=0.18\textwidth]{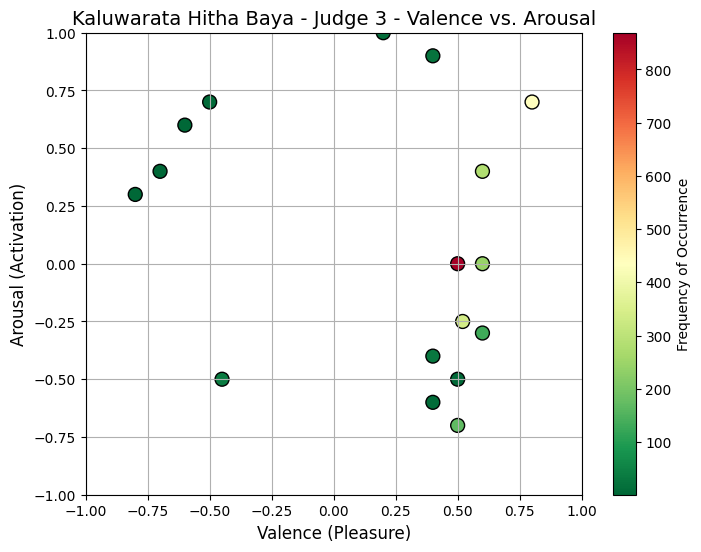} &

    \includegraphics[width=0.18\textwidth]{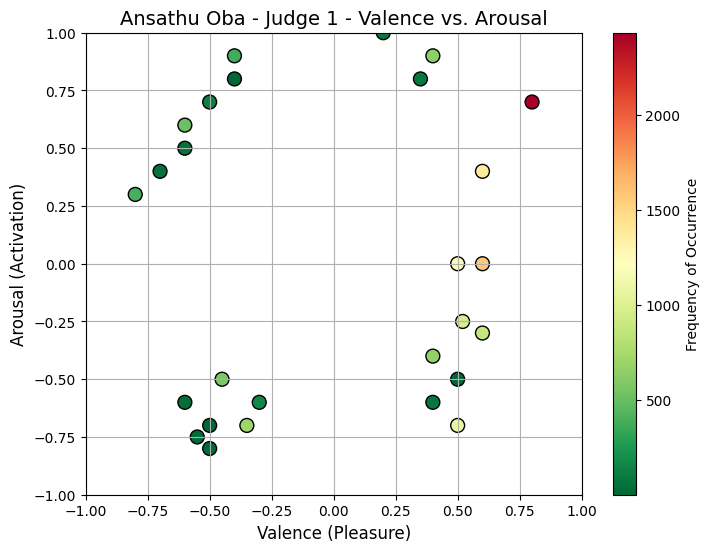} & 
    \includegraphics[width=0.18\textwidth]{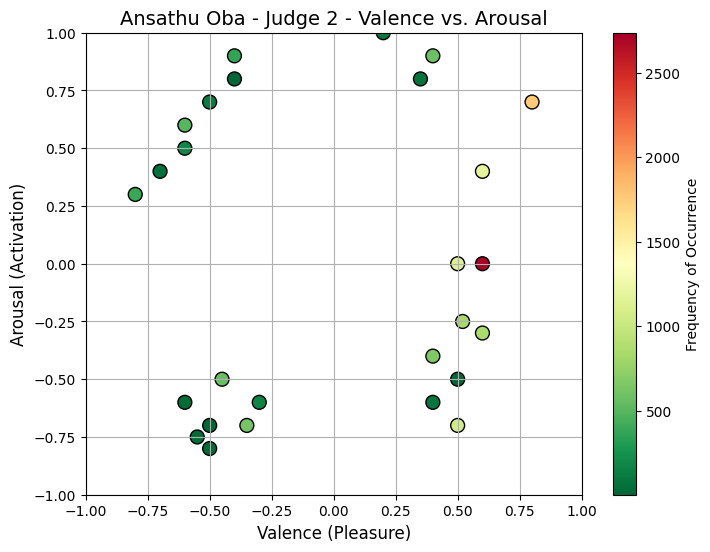} & 
    \includegraphics[width=0.18\textwidth]{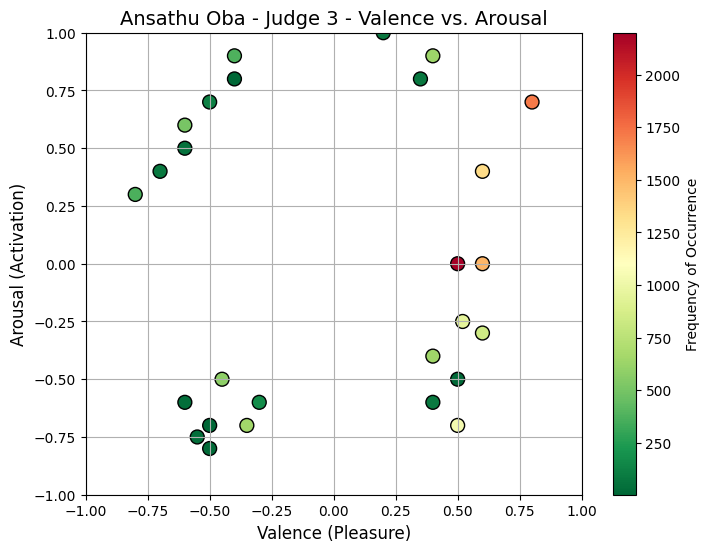} \\

\end{tabular}
}
\caption{Frequency distribution of emotions annotated by three independent judges. Each row represents a song, with the columns showing the emotional frequency derived from comments for each annotator.}
\label{fig1}
\end{figure}

The comment mapping in this study is based on Russell’s emotion model~\cite{russell1980circumplex}, which categorizes emotions using valence and arousal values. This method allowed for a clear visualization of the emotional intensity associated with each song in Figure~\ref{fig1}, the frequency of emotions assigned to comments corresponding to a song is visually represented through a color gradient, where a lower frequency of emotions is depicted in green, while a higher frequency is represented in red.
During the annotation process, out of a total of 63,471 comments, 56 comments were identified as unclassifiable by at least two annotators, with the majority of these cases being unanimously unclassifiable by all three annotators. All remaining comments were categorized into one of the emotions provided under Russell's matrix.
To evaluate the reliability of the annotation process,  \textit{Fleiss' kappa} measure was employed, a statistical metric used to evaluate inter-annotator agreement in categorical data. The results indicated a high level of agreement among the annotators, with a \textit{Fleiss' kappa} score of 84.96\%. This substantial level of agreement demonstrates the consistency and reliability of the annotation process, reinforcing the validity of the data set for subsequent analysis.
Following the annotation process,  distribution of emotions across the dataset was analyzed, examining the frequency of each emotion within comments associated with the twenty selected songs. This analysis provided insights into the predominant emotional expressions elicited by different songs. The graphical representation of the emotional distributions is presented in Figures~\ref{fig1} illustrating the varying emotional intensities across the annotated data set. This figure further illustrates how well the three annotators are in agreement,t justifying the above value received for the \textit{Fleiss’ kappa} score.

Analysis of the distribution of comments across the four quadrants of~\citet{russell1980circumplex}'s emotion model indicates that the comments for all songs exhibit a spread throughout all quadrants, demonstrating a diverse pattern of emotional responses. However, when considering the highest emotion that occurs for each song, clear trends emerge. For the benefit of international readers, each of the songs is given with a song index \texttt{Sx} where \texttt{x} is a number.  The index mapping to the song title is provided in Table~\ref{tab:emotion_summary} along with the total comment counts. Further, the emotions reported at a maximum and minimum frequency for the comments of each of the songs were shown along with the relevant count.
These results highlight how comments on different songs display varying emotional responses, reinforcing the effectiveness of emotion mapping in understanding listener engagement and sentiment distribution.

\begin{table}[!htb]
    \centering
    \caption{Total Comment Count as Well as the Maximum and Minimum Emotions Reported for Comments of Each Song}
    \label{tab:emotion_summary}
   \includegraphics[width=\textwidth]{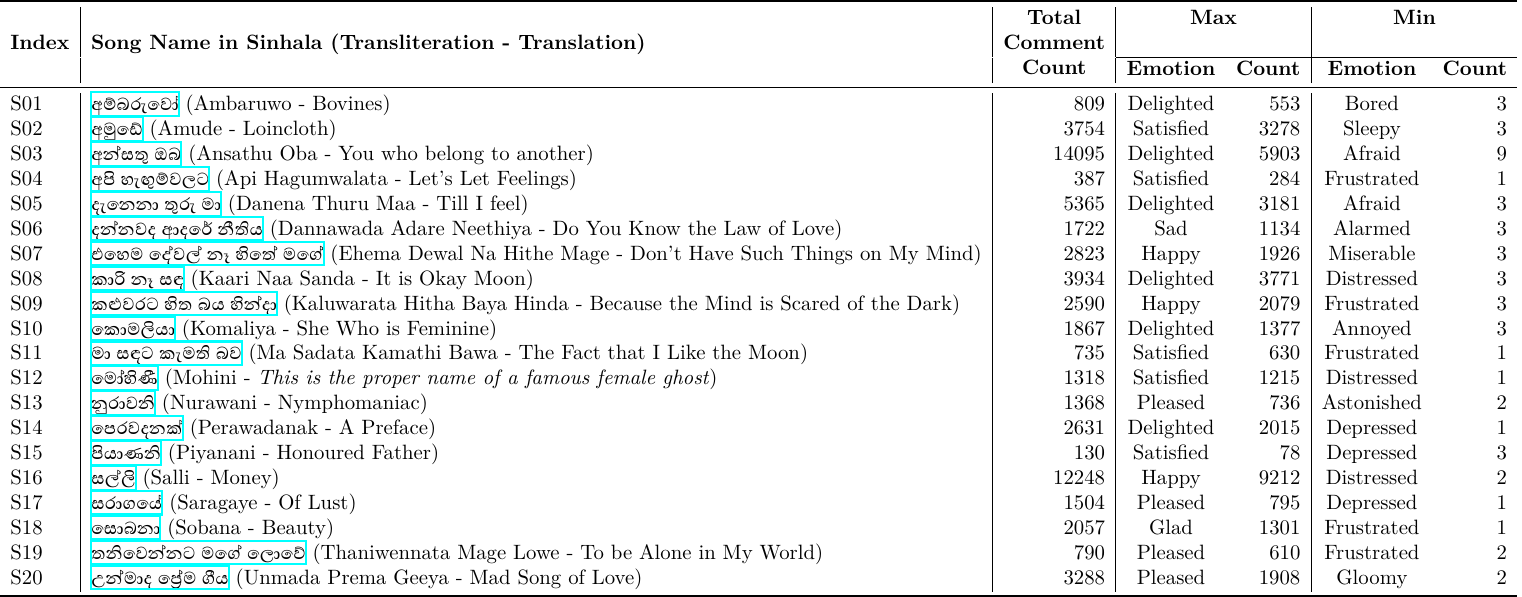}
    
    
\end{table}

Several songs exhibit the highest frequency of comments associated with emotions in Quadrant 1, which corresponds to high arousal and high valence. These include \texttt{S01}, \texttt{S03}, \texttt{S05}, \texttt{S07}, \texttt{S08}, \texttt{S09}, \texttt{S10}, \texttt{S13}, \texttt{S14}, \texttt{S16}, \texttt{S17}, \texttt{S18}, \texttt{S19}, and \texttt{S20}. These songs elicit strong emotional responses, often associated with excitement, joy, or enthusiasm.
In contrast, \texttt{S06} is the only song that exhibits the highest emotional frequency in Quadrant 3, which is characterized by low arousal and low valence. This suggests that the song evokes subdued, melancholic, or even depressive emotional responses.

A different trend is observed in Quadrant 2, which is associated with positive but low-energy emotions such as calmness and contentment. The songs \texttt{S02}, \texttt{S04}, \texttt{S11}, \texttt{S12}, and \texttt{S15} have the highest frequency of emotions mapped to this quadrant, suggesting that they generate a more soothing and peaceful emotional response among listeners.
Conversely, when considering the lowest frequency of emotional occurrences,  \texttt{S03}, \texttt{S04}, \texttt{S05}, \texttt{S06}, \texttt{S08}, \texttt{S09}, \texttt{S10}, \texttt{S11}, \texttt{S12}, \texttt{S16}, \texttt{S18}, and \texttt{S19} exhibit the least frequent emotional responses in Quadrant 4, which represents low valence and high arousal, often associated with distress, anxiety, or unease. Additionally, \texttt{S01}, \texttt{S07}, \texttt{S14}, \texttt{S15}, \texttt{S17}, and \texttt{S20} have the lowest emotional frequency in Quadrant 3, meaning they rarely evoke sadness or low-energy negative emotions. The song \texttt{S02} has the lowest occurring emotion in Quadrant 2, indicating that it is not frequently associated with calm or peaceful emotions. On the other hand, \texttt{S13} exhibits the lowest frequency of emotional responses in Quadrant 1, suggesting that it is less likely to generate intense positive emotions such as excitement or joy.

\subsection{Comparison of Comment-Based and Song-Based Emotion Categorization}

\begin{table}[!htb]
\caption{Summary of Emotion Analysis for Songs}
\label{tab:emotion_summary2}
\centering
\resizebox{0.9\textwidth}{!}{
\begin{tabular}{l|c|c|c}
\toprule
\textbf{Song} & \makecell{\textbf{Emotion Based on Song}\\\textbf{Classification}}
 & \makecell{\textbf{Highest Frequency Emotion}\\\textbf{Based on Comments}} & \makecell{\textbf{Cosine }\\\textbf{Similarity}} \\
\midrule
\texttt{S01} & Relaxed & Delighted & -0.098 \\
\texttt{S02} & Sad & Satisfied & -0.202 \\
\texttt{S03} & Tense & Delighted & -0.067 \\
\texttt{S04} & Depressed & Satisfied & 0.065 \\
\texttt{S05} & Distressed & Delighted & -0.157 \\
\texttt{S06} & Frustrated & Sad & 0.212 \\
\texttt{S07} & Relaxed & Happy & 0.581 \\
\texttt{S08} & Sad & Delighted & -0.993 \\
\texttt{S09} & Afraid & Happy & -0.447 \\
\texttt{S10} & Happy & Delighted & 0.753 \\
\texttt{S11} & Angry & Satisfied & -0.994 \\
\texttt{S12} & Satisfied & Satisfied & 1.000 \\
\texttt{S13} & Calm & Pleased & 0.707 \\
\texttt{S14} & Sad & Delighted & -0.993 \\
\texttt{S15} & Sad & Satisfied & -0.202 \\
\texttt{S16} & Delighted & Happy & 0.753 \\
\texttt{S17} & Relaxed & Pleased & 0.581 \\
\texttt{S18} & Relaxed & Glad & 0.032 \\
\texttt{S19} & Sad & Pleased & -0.354 \\
\texttt{S20} & Happy & Pleased & 1.000 \\
\bottomrule
\end{tabular}
}
\end{table}

\begin{figure}[!htb]
    \centering
    \caption{Emotion visualization for Sinhala songs based on comment analysis and annotation }
    \label{fig:emotion_visualization_1}
    
    \begin{minipage}{0.24\linewidth}
        \centering
        \includegraphics[width=\linewidth]{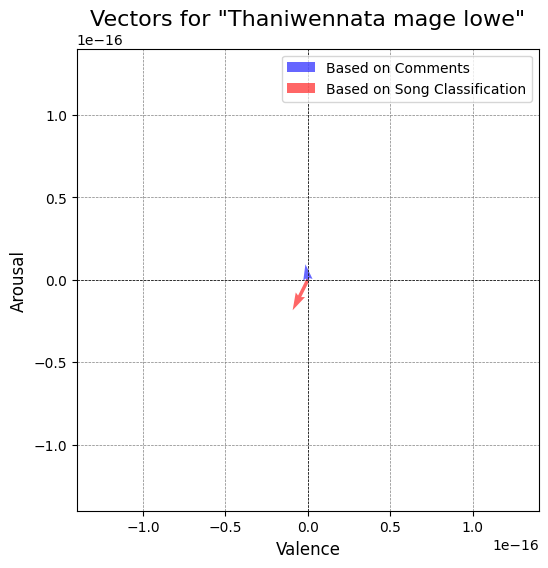}
        \includegraphics[width=\linewidth]{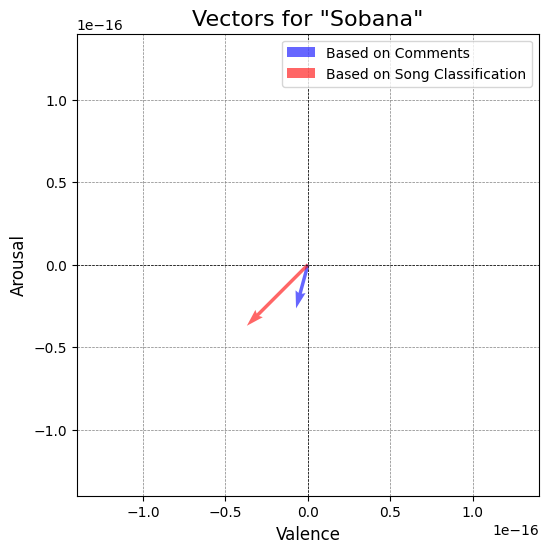}
        \includegraphics[width=\linewidth]{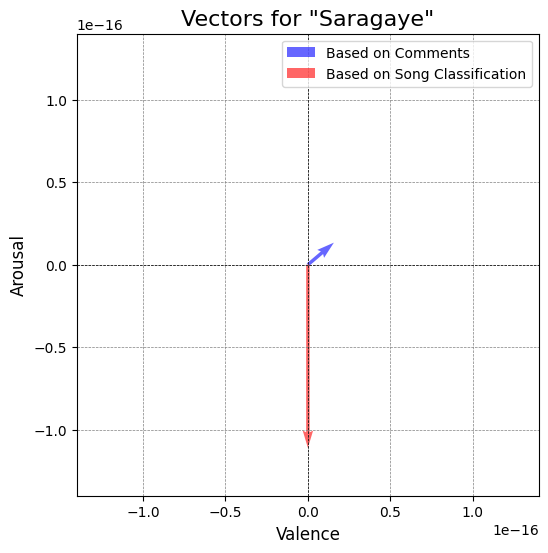}
        \includegraphics[width=\linewidth]{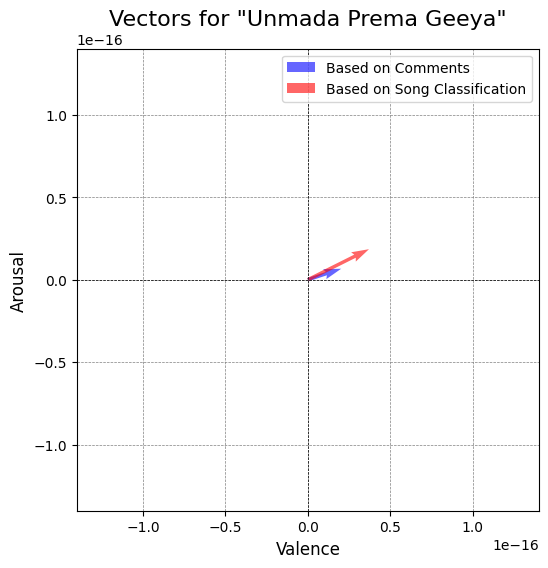}
        \includegraphics[width=\linewidth]{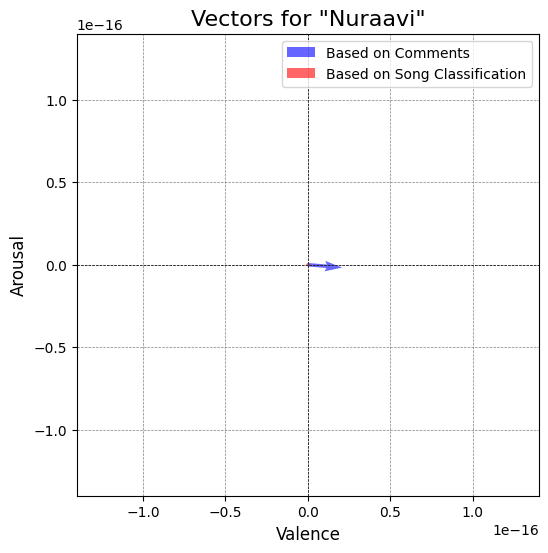}
    \end{minipage}
    \begin{minipage}{0.24\linewidth}
        \centering
        \includegraphics[width=\linewidth]{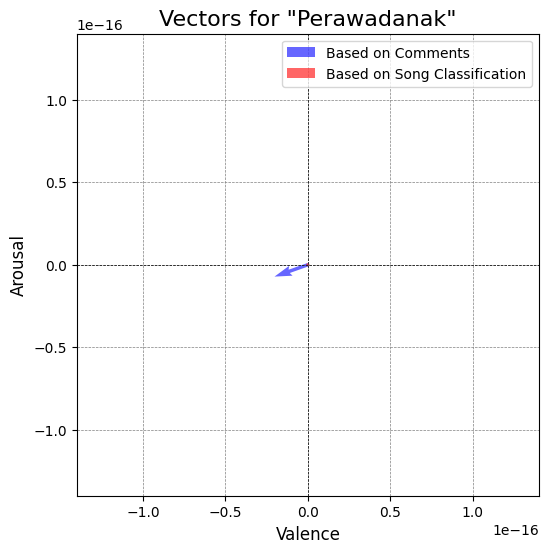}
        \includegraphics[width=\linewidth]{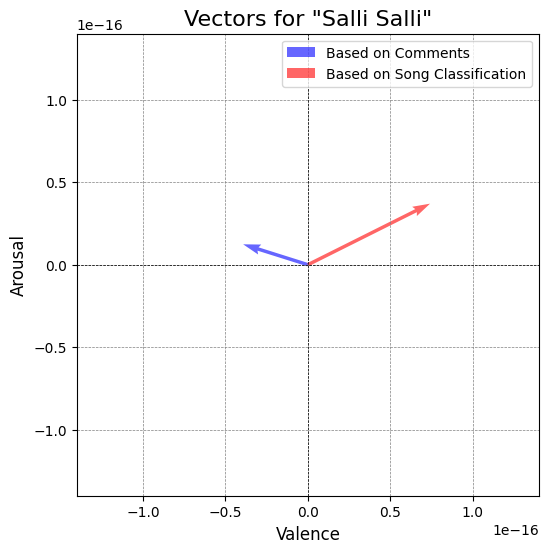}
        \includegraphics[width=\linewidth]{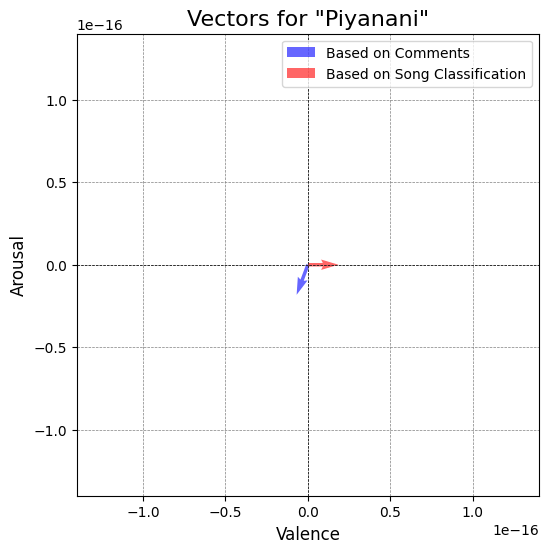}
        \includegraphics[width=\linewidth]{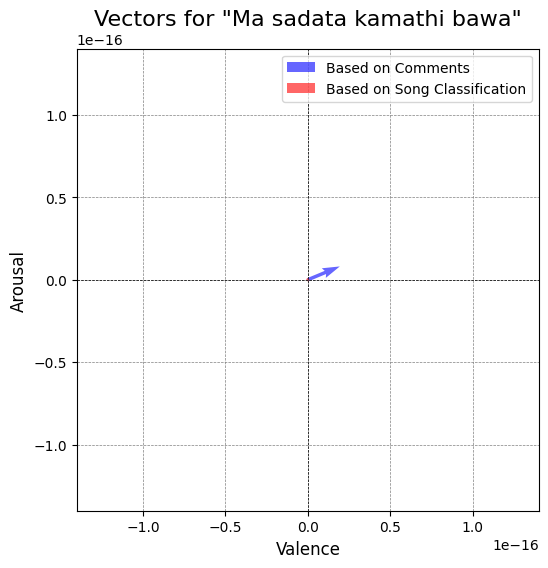}
        \includegraphics[width=\linewidth]{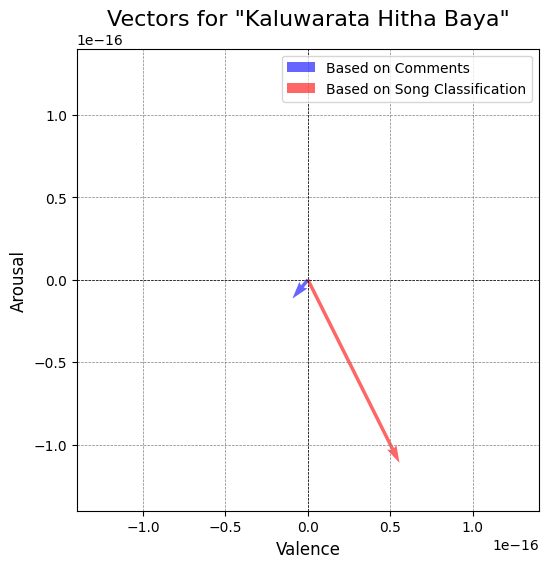}
    \end{minipage}
    \begin{minipage}{0.24\linewidth}
        \centering
        \includegraphics[width=\linewidth]{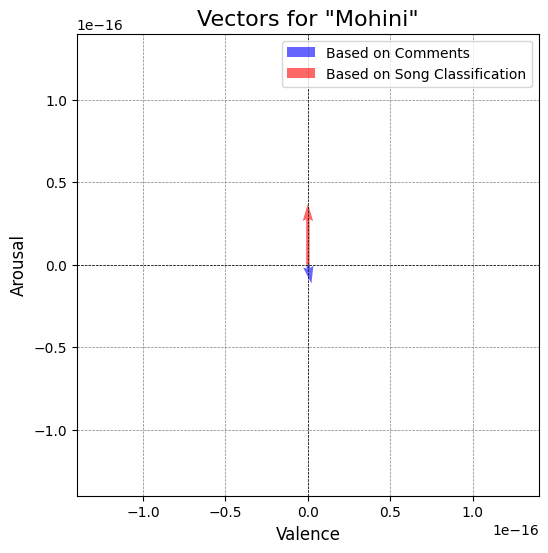}
        \includegraphics[width=\linewidth]{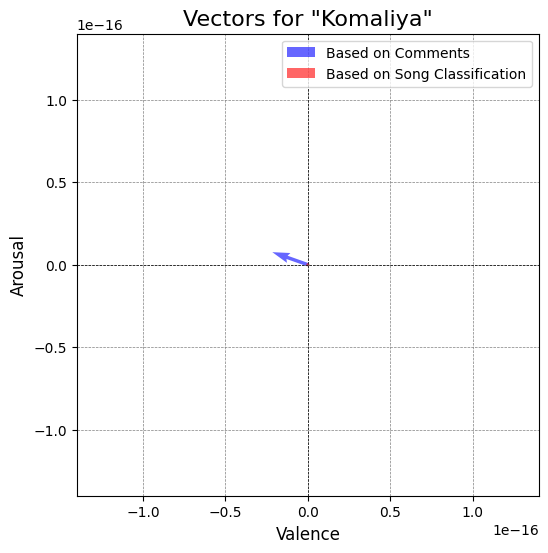}
        \includegraphics[width=\linewidth]{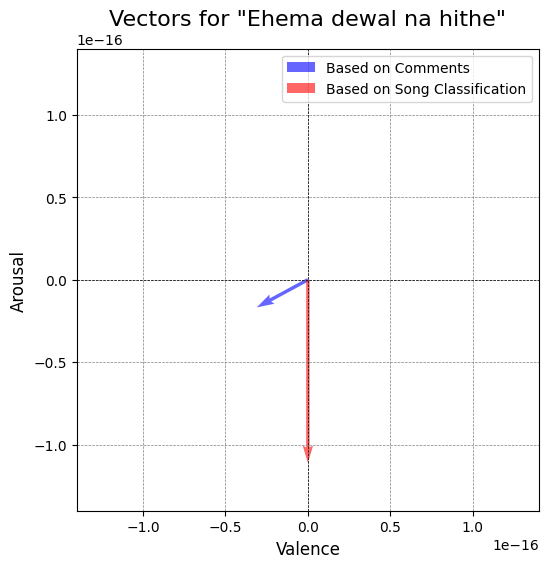}
        \includegraphics[width=\linewidth]{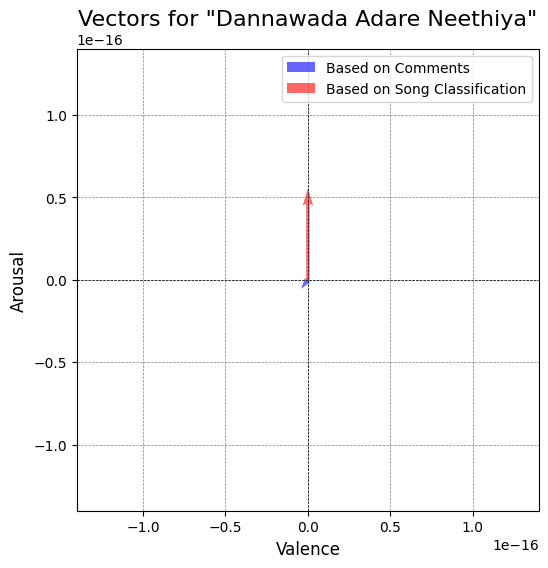}
        \includegraphics[width=\linewidth]{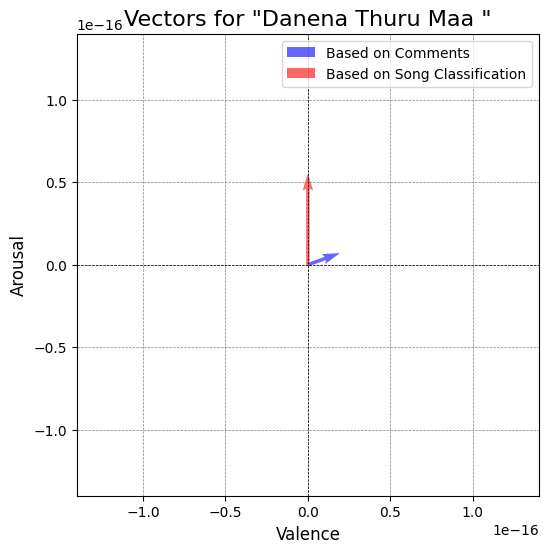}
    \end{minipage}
     \begin{minipage}{0.24\linewidth}
        \centering
        \includegraphics[width=\linewidth]{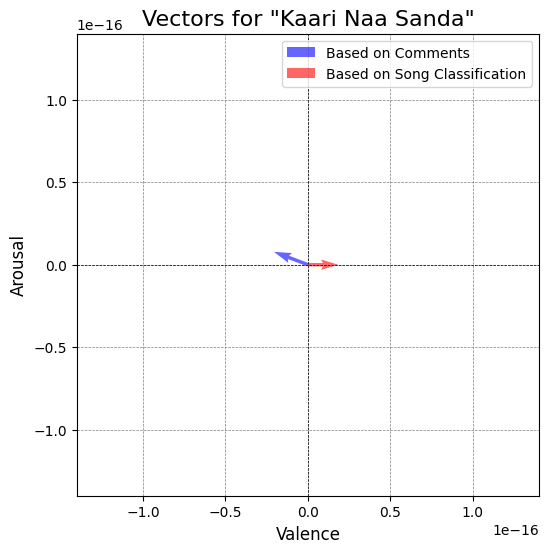}
        \includegraphics[width=\linewidth]{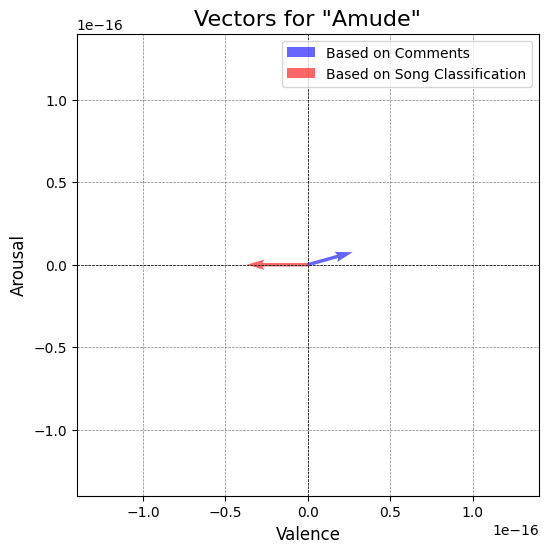}
        \includegraphics[width=\linewidth]{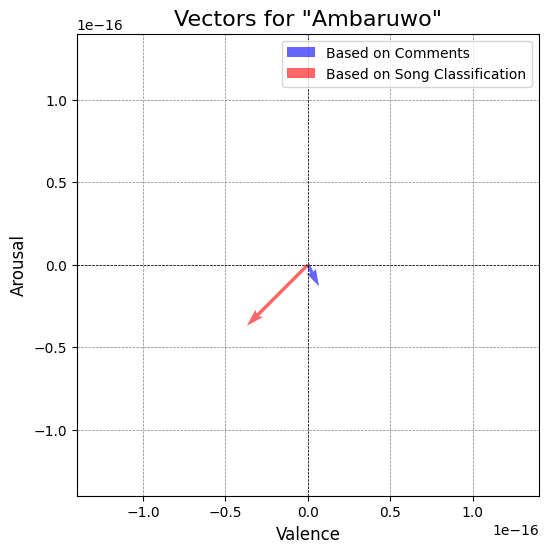}
        \includegraphics[width=\linewidth]{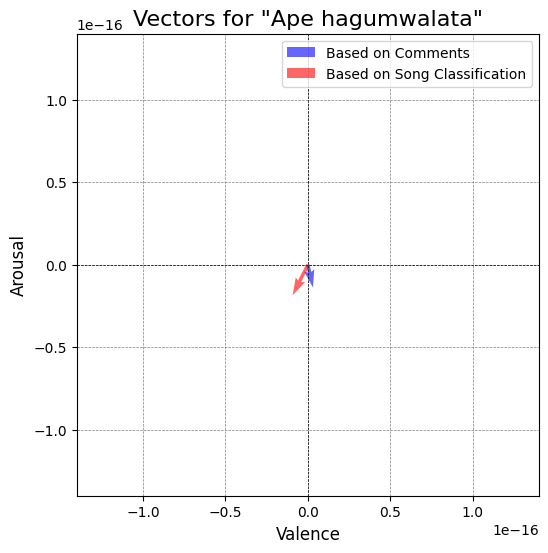}
        \includegraphics[width=\linewidth]{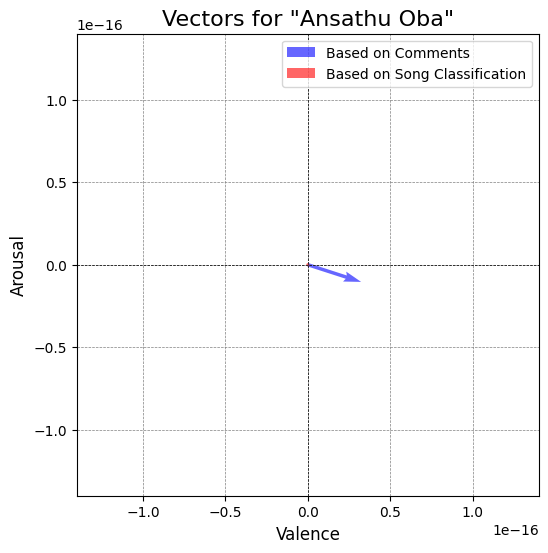}
    \end{minipage}   
\end{figure}

To evaluate the relationship between the emotional responses reflected in the YouTube comments corresponding to a song and the manually annotated song emotions, the highest frequency emotion vector derived from the comments with the dominant emotion assigned to the song itself by the annotators were compared. This comparison was performed using cosine similarity, a widely used metric for measuring the alignment between two vectors in a multi-dimensional space. The similarity values obtained from this comparison are presented in Table~\ref{tab:emotion_summary2}.
As discussed in Section~\ref{sec:meth:son}, the direct comparison of Table~\ref{tab:emotion_summary2} was deemed in adequate and a more extensive comparison was conducted using the normalized emotion vectors. This relationship between the emotion vectors based on standardized comments and the emotion vectors derived from the categorization based on songs is visualized in Figure~\ref{fig:emotion_visualization_1} which effectively illustrates the degree of alignment and divergence between the two sources of emotional classification, providing valuable insights into how listener perceptions correspond to expert annotations of emotional expression in Sinhala songs.

\subsection{Feature Selection and Model Pre-Training using the Sinhala News Comment Classification Baseline}

\newcommand{\R}[3][]{\makecell{#2$\pm$#3\ifthenelse{\isempty{#1}}{}{\\(#1)}}}

\begin{table}[!htbp]
\caption{Classification Results for Different Models and Feature Representations on the Sinhala News Classification Task. ** Results given in parentheses are the results reported in study~\citet{ranathunga2021sentiment}. All other results are our own experiments.}
    \centering
    \scriptsize
    \renewcommand{\arraystretch}{1.2}
    \begin{adjustbox}{max width=\textwidth}
    \begin{tabular}{lcccccccccccccccc}
        \toprule
        \multirow{2}{*}{\textbf{Feature}} 
            & \multicolumn{4}{c}{\textbf{Accuracy}} 
            & \multicolumn{4}{c}{\textbf{Precision}} 
            & \multicolumn{4}{c}{\textbf{Recall}} 
            & \multicolumn{4}{c}{\textbf{F1 Score}} \\
        \cmidrule(lr){2-5} \cmidrule(lr){6-9} \cmidrule(lr){10-13} \cmidrule(lr){14-17}
        & LR & RF & SVM & XGBoost 
        & LR & RF & SVM & XGBoost 
        & LR & RF & SVM & XGBoost 
        & LR & RF & SVM & XGBoost \\
        \midrule

        TF-IDF 
        & \R[0.85]{0.743}{0.003} 
        & \R{0.727}{0.007} 
        & \R[0.85]{0.745}{0.008} 
        & \R{0.751}{0.009} 
        & \R[0.9]{0.743}{0.003} 
        & \R{0.738}{0.008} 
        & \R[0.89]{0.745}{0.008} 
        & \R{0.753}{0.010} 
        & \R[0.78]{0.743}{0.003} 
        & \R{0.727}{0.007} 
        & \R[0.81]{0.745}{0.008} 
        & \R{0.751}{0.009} 
        & \R[0.84]{0.743}{0.003} 
        & \R{0.724}{0.007} 
        & \R[0.85]{0.744}{0.008} 
        & \R{0.751}{0.009} \\

        Word2Vec 
        & \R{0.640}{0.143} 
        & \R{0.630}{0.127} 
        & \R{0.638}{0.141} 
        & \R{0.638}{0.135} 
        & \R{0.516}{0.269} 
        & \R{0.505}{0.252} 
        & \R{0.514}{0.267} 
        & \R{0.514}{0.261} 
        & \R{0.640}{0.143} 
        & \R{0.630}{0.127} 
        & \R{0.638}{0.141} 
        & \R{0.638}{0.135} 
        & \R{0.556}{0.226} 
        & \R{0.547}{0.210} 
        & \R{0.554}{0.224} 
        & \R{0.554}{0.217} \\

        fastText 
        & \R[0.88]{0.601}{0.098} 
        & \R{0.599}{0.096} 
        & \R[0.88]{0.597}{0.100} 
        & \R{0.611}{0.108} 
        & \R[0.9]{0.476}{0.223} 
        & \R{0.474}{0.221} 
        & \R[0.89]{0.472}{0.225} 
        & \R{0.486}{0.233} 
        & \R[0.85]{0.601}{0.098} 
        & \R{0.599}{0.096} 
        & \R[0.86]{0.597}{0.100} 
        & \R{0.611}{0.108} 
        & \R[0.88]{0.517}{0.181} 
        & \R{0.515}{0.179} 
        & \R[0.87]{0.513}{0.183} 
        & \R{0.528}{0.191}\\

        SBERT 
        & \R{0.555}{0.058} 
        & \R{0.555}{0.052} 
        & \R{0.556}{0.059} 
        & \R{0.554}{0.051} 
        & \R{0.432}{0.184} 
        & \R{0.432}{0.179} 
        & \R{0.433}{0.186} 
        & \R{0.430}{0.177} 
        & \R{0.555}{0.058} 
        & \R{0.555}{0.052} 
        & \R{0.556}{0.059} 
        & \R{0.554}{0.051} 
        & \R{0.471}{0.141} 
        & \R{0.469}{0.132} 
        & \R{0.471}{0.141} 
        & \R{0.470}{0.133} \\

        SinBERT 
        & \R{0.794}{0.006} 
        & \R{0.765}{0.008} 
        & \R{0.789}{0.008} 
        & \R{0.800}{0.007} 
        & \R{0.794}{0.007} 
        & \R{0.766}{0.008} 
        & \R{0.789}{0.008} 
        & \R{0.804}{0.008} 
        & \R{0.794}{0.006} 
        & \R{0.765}{0.008} 
        & \R{0.789}{0.008} 
        & \R{0.800}{0.007} 
        & \R{0.794}{0.006} 
        & \R{0.765}{0.008} 
        & \R{0.789}{0.008} 
        & \R{0.800}{0.007} \\

        \midrule
         
        Word2Vec + fastText
        & \R[0.89]{0.637}{0.140} 
        & \R{0.627}{0.124} 
        & \R{0.633}{0.136} 
        & \R{0.633}{0.130} 
        & \R[0.9]{0.512}{0.265} 
        & \R{0.502}{0.249} 
        & \R{0.509}{0.262} 
        & \R{0.510}{0.257} 
        & \R[0.85]{0.637}{0.140} 
        & \R{0.627}{0.124} 
        & \R{0.633}{0.136} 
        & \R{0.633}{0.130} 
        & \R[0.87]{0.553}{0.223} 
        & \R{0.544}{0.207} 
        & \R{0.550}{0.220} 
        & \R{0.549}{0.213} \\

        Word2Vec + SBERT 
        & \R{0.639}{0.136} 
        & \R{0.621}{0.118} 
        & \R{0.640}{0.143} 
        & \R{0.626}{0.123} 
        & \R{0.515}{0.262} 
        & \R{0.500}{0.247} 
        & \R{0.515}{0.268} 
        & \R{0.502}{0.249} 
        & \R{0.639}{0.136} 
        & \R{0.621}{0.118} 
        & \R{0.640}{0.143} 
        & \R{0.626}{0.123} 
        & \R{0.556}{0.219} 
        & \R{0.537}{0.200} 
        & \R{0.556}{0.226} 
        & \R{0.542}{0.206} \\

        Word2Vec + SinBERT 
        & \R{\textbf{0.801}}{\textbf{0.004}} 
        & \R{0.769}{0.010} 
        & \R{0.795}{0.007} 
        & \R{0.803}{0.005} 
        & \R{\textbf{0.801}}{\textbf{0.004}} 
        & \R{0.770}{0.010} 
        & \R{0.795}{0.007} 
        & \R{0.807}{0.006} 
        & \R{\textbf{0.801}}{\textbf{0.004}} 
        & \R{0.769}{0.010} 
        & \R{0.795}{0.007} 
        & \R{0.803}{0.005} 
        & \R{\textbf{0.801}}{\textbf{0.004}} 
        & \R{0.769}{0.010} 
        & \R{0.795}{0.007} 
        & \R{0.802}{0.005} \\
        
        fastText + SBERT 
        & \R{0.605}{0.108} 
        & \R{0.593}{0.090} 
        & \R{0.599}{0.102} 
        & \R{0.604}{0.101} 
        & \R{0.480}{0.233} 
        & \R{0.471}{0.218} 
        & \R{0.474}{0.227} 
        & \R{0.480}{0.227} 
        & \R{0.605}{0.108} 
        & \R{0.593}{0.090} 
        & \R{0.599}{0.102} 
        & \R{0.604}{0.101} 
        & \R{0.521}{0.191} 
        & \R{0.508}{0.171} 
        & \R{0.516}{0.186} 
        & \R{0.521}{0.184} \\
        
        fastText + SinBERT 
        & \R{0.794}{0.007} 
        & \R{\textbf{0.772}}{\textbf{0.014}} 
        & \R{0.789}{0.009} 
        & \R{0.804}{0.006} 
        & \R{0.795}{0.007} 
        & \R{\textbf{0.773}}{\textbf{0.013}} 
        & \R{0.790}{0.009} 
        & \R{0.807}{0.006} 
        & \R{0.794}{0.007} 
        & \R{\textbf{0.772}}{\textbf{0.014}} 
        & \R{0.789}{0.009} 
        & \R{0.804}{0.006} 
        & \R{0.794}{0.007} 
        & \R{\textbf{0.772}}{\textbf{0.014}} 
        & \R{0.789}{0.009} 
        & \R{0.803}{0.006} \\

        SBERT + SinBERT 
        & \R{0.794}{0.006} 
        & \R{0.767}{0.008} 
        & \R{0.788}{0.008} 
        & \R{0.799}{0.007} 
        & \R{0.795}{0.006} 
        & \R{0.767}{0.008} 
        & \R{0.788}{0.008} 
        & \R{0.804}{0.007} 
        & \R{0.794}{0.006} 
        & \R{0.767}{0.008} 
        & \R{0.788}{0.008} 
        & \R{0.799}{0.007} 
        & \R{0.794}{0.006} 
        & \R{0.767}{0.008} 
        & \R{0.788}{0.008} 
        & \R{0.798}{0.007} \\

        \midrule

        Word2Vec + fastText + SBERT 
        & \R{0.635}{0.132} 
        & \R{0.629}{0.126} 
        & \R{0.627}{0.130} 
        & \R{0.633}{0.130} 
        & \R{0.510}{0.257} 
        & \R{0.506}{0.253} 
        & \R{0.503}{0.256} 
        & \R{0.510}{0.257} 
        & \R{0.635}{0.132} 
        & \R{0.629}{0.126} 
        & \R{0.627}{0.130} 
        & \R{0.633}{0.130} 
        & \R{0.551}{0.215} 
        & \R{0.546}{0.209} 
        & \R{0.544}{0.214} 
        & \R{0.550}{0.213} \\

        Word2Vec + fastText + SinBERT 
        & \R{0.797}{0.006} 
        & \R{0.771}{0.007} 
        & \R{0.786}{0.008} 
        & \R{\textbf{0.809}}{\textbf{0.007}} 
        & \R{0.798}{0.005} 
        & \R{0.772}{0.007} 
        & \R{0.787}{0.008} 
        & \R{\textbf{0.810}}{\textbf{0.007}} 
        & \R{0.797}{0.006} 
        & \R{0.771}{0.007} 
        & \R{0.786}{0.008} 
        & \R{\textbf{0.809}}{\textbf{0.007}} 
        & \R{0.797}{0.006} 
        & \R{0.771}{0.007} 
        & \R{0.786}{0.008} 
        & \R{\textbf{0.808}}{\textbf{0.008}} \\
          
        Word2Vec + SBERT + SinBERT 
        & \R{0.798}{0.007} 
        & \R{0.766}{0.008} 
        & \R{\textbf{0.795}}{\textbf{0.009}} 
        & \R{0.799}{0.008} 
        & \R{0.799}{0.007} 
        & \R{0.766}{0.008} 
        & \R{\textbf{0.796}}{\textbf{0.009}} 
        & \R{0.801}{0.008} 
        & \R{0.798}{0.007} 
        & \R{0.766}{0.008} 
        & \R{\textbf{0.795}}{\textbf{0.009}} 
        & \R{0.799}{0.008} 
        & \R{0.798}{0.007} 
        & \R{0.766}{0.008} 
        & \R{\textbf{0.795}}{\textbf{0.009}} 
        & \R{0.798}{0.008} \\

        fastText + SBERT + SinBERT
        & \R{0.792}{0.012} 
        & \R{0.769}{0.011} 
        & \R{0.792}{0.012} 
        & \R{0.797}{0.006} 
        & \R{0.793}{0.012} 
        & \R{0.770}{0.010} 
        & \R{0.792}{0.012} 
        & \R{0.800}{0.006} 
        & \R{0.792}{0.012} 
        & \R{0.769}{0.011} 
        & \R{0.792}{0.012} 
        & \R{0.797}{0.006} 
        & \R{0.792}{0.012} 
        & \R{0.769}{0.011} 
        & \R{0.792}{0.012} 
        & \R{0.797}{0.006} \\
        
       \midrule    
        
        Word2Vec + fastText + SinBERT + SBERT 
        & \R{0.792}{0.007} 
        & \R{0.768}{0.008} 
        & \R{0.787}{0.008} 
        & \R{0.800}{0.010} 
        & \R{0.793}{0.007} 
        & \R{0.769}{0.008} 
        & \R{0.788}{0.008} 
        & \R{0.801}{0.010} 
        & \R{0.792}{0.007} 
        & \R{0.768}{0.008} 
        & \R{0.787}{0.008} 
        & \R{0.800}{0.010} 
        & \R{0.792}{0.007} 
        & \R{0.768}{0.008} 
        & \R{0.787}{0.008} 
        & \R{0.800}{0.010} \\

        \bottomrule
    \end{tabular}
    \end{adjustbox}
    
    \label{tab:classification_results}
\end{table}

Given that the work by~\citet{ranathunga2021sentiment} was the closest to our task, and their task includes a large Sinhala comment classification data set, this study re-implemented their methodology so that it can do feature selection and pre-train our classification model. Given that it is a re-implementation and an extension, the models are trained on the same Sinhala news classification task on which they have reported the results. The result of these experiments are shown in Table~\ref{tab:classification_results}. This study extend the experiments in both the feature direction and the model direction. In the model direction, experiments of Random Forest (RF) and XGBoost were added to their existing Logistic Regression and Support Vector Machine models. This study introduce SBERT and SinBERT features as well as experiment with some future combinations that they have not tried.   

The original study by~\citet{ranathunga2021sentiment} reported strong classification performance, with \textit{F1-scores} reaching up to 0.88 using Logistic Regression with fastText embeddings. However, our replication study was unable to reach these values. Given that this may be due to the random splits of Training and Test sets, the experiment was run 10 times and report the median with the margin of error. 
Nevertheless, the results highlight how 
various feature combinations influenced the performance of the classifiers, demonstrating that certain feature sets
performed comparatively well.
In particular, XGBoost with \texttt{Word2Vec+fastText+SinBERT} achieved the highest \textit{accuracy} of 0.808, suggesting that combining multiple embedding types can enhance classification accuracy. 
%

Then, deep learning models were tested on the same news comment classification task using \texttt{Word2Vec+fastText+SinBERT} features.
%
The Multi-Layer Perceptron (MLP) model was trained on a flattened version of the embeddings, while the LSTM and CNN models processed the embeddings as sequential inputs. 
The best results for each model were recorded in Table~\ref{tab:performance_deeplearn}.

\begin{table}[!htb]
 \caption{Performance Metrics of Deep Learning Models}
    \centering
    \resizebox{0.8\textwidth}{!}{
    \begin{tabular}{c|c|c|c|c|c}
        \hline
        Model & Accuracy & Precision & Recall & F1 Score & ROC AUC \\
        \hline
        MLP  & 0.837 & 0.850 & 0.792 & \textbf{0.820} & \textbf{0.836} \\
        LSTM & 0.759 & 0.780 & 0.678 & 0.726 & 0.754 \\
        CNN  & 0.785 & 0.798 & 0.727 & 0.761 & 0.782 \\
        \hline
    \end{tabular}
}   
    \label{tab:performance_deeplearn}
\end{table}

The MLP model achieved the highest performance in all evaluation metrics, even surpassing the results of ML models shown in Table~\ref{tab:classification_results}.  
This could be attributed to its ability to capture global dependencies within feature representations more efficiently than sequential models such as LSTM, which require longer training durations to capture temporal dependencies. Although CNN and LSTM models leveraged spatial and sequential structures, their performance was slightly hindered, likely due to the nature of the data set where dense embeddings without explicit sequential relationships played a crucial role in classification. 
For model evaluation, primarily focus was on the \textit{ROC-AUC score}, which measures the trade-off between the true positive rate and the false positive rate.
Therefore, based on the experimental findings, the MLP model was determined to be the best performing approach for this task with excellent \textit{ROC-AUC score} as well as comparable results to XGBoost from the classical ML model experiment F1 score was higher.

\begin{table}[!htb]
\centering
\caption{Performance of top five MLP models}
\label{tab:mlp_opsperformance}
\resizebox{0.8\textwidth}{!}{
\begin{tabular}{D{.}{.}{3}|c|c|D{.}{.}{4}}
\toprule
\multicolumn{1}{c|}{ROC AUC} & Layer Sizes & Dropout Rates & \multicolumn{1}{c}{Learning Rate} \\
\midrule
0.836 &  [512, 256] & [0.3, 0.2] & 0.001 \\
0.881 &  [512, 256] & [0.3, 0.2] & 0.001 \\
0.882 &  [128, 64] & [0.2, 0.1] & 0.001 \\
0.882 &  [512, 256, 128] & [0.4, 0.3, 0.2] & 0.0005 \\
0.887 &  [256, 128, 64] & [0.3, 0.2, 0.1] & 0.0003 \\
\bottomrule
\end{tabular}
}
\end{table}

\begin{table*}[!htbp]
    \small
    \renewcommand{\arraystretch}{1.1}
    \setlength{\tabcolsep}{4pt} 
    
    \caption{Emotion classification using the pre-trained MLP model on YouTube Sinhala Comments. The \textbf{Total} column lists the total number of comments tagged under the relevant emotion by a judge. The \textbf{MLP Result} column then gives the number of those comments classified by the MLP as \texttt{POSITIVE}~\TikzPos{} and \texttt{NEGATIVE}~\TikzNeg{}, respectively. Note that in Russell's Valence-Arousal model, High-Valence corresponds to positivity. Note that this table only lists emotions that had non-zero counts to avoid wasting space.}
    \label{tab:mlp_classification}
\resizebox{\textwidth}{!}{    
    \begin{tabular}{c|l|r|c|r|c|r|c}
        \toprule
        \multicolumn{2}{c|}{\multirow{2}{*}{Emotion}} & \multicolumn{2}{c|}{Judge 1} & \multicolumn{2}{c|}{Judge 2} & \multicolumn{2}{c}{Judge 3} \\
        \hhline{~~------}
         \multicolumn{2}{c|}{} & Total & MLP Result & Total & MLP Result & Total & MLP Result \\
        \midrule
        \multirow{7}{*}[-2.4ex]{\rotatebox[origin=c]{90}{\scriptsize High-Valance High-Arousal}} & Aroused     & 602 & \stackedbar{437}{165}   & 558 & \stackedbar{397}{161}    & 573 & \stackedbar{414}{159}    \\
        & Astonished  & 815 & \stackedbar{532}{283}  & 762  & \stackedbar{501}{261}    & 808 & \stackedbar{530}{278}    \\        
        & Delighted   & 11789 & \stackedbar{7407}{4382}  & 11323 & \stackedbar{7131}{4192}  & 12464 & \stackedbar{7951}{4513}   \\   
        & Excited     & 1735 & \stackedbar{1205}{530}   & 1640  & \stackedbar{1151}{489}   & 2672 & \stackedbar{2033}{639}    \\    
        & Glad        & 5969 & \stackedbar{3704}{2265}  & 6200  & \stackedbar{3986}{2214}  & 5984 & \stackedbar{3740}{2244}    \\   
        & Happy       & 11391 & \stackedbar{8112}{3279}  & 10090 & \stackedbar{6972}{3118}  & 11590 & \stackedbar{8180}{3410}   \\   
        & Pleased     & 7330 & \stackedbar{5088}{2242}  & 8539  & \stackedbar{5900}{2639}  & 6483 & \stackedbar{4491}{1992}   \\    \hline    
        \multirow{6}{*}{\rotatebox[origin=c]{90}{\scriptsize High-Valance Low-Arousal}} & Calm        & 2315 & \stackedbar{1626}{689}   & 2259  & \stackedbar{1597}{662}  & 2264  & \stackedbar{1586}{678}    \\    
        & Content     & 3763 & \stackedbar{2567}{1196}  & 3729  & \stackedbar{2555}{1174}  & 3443  & \stackedbar{2348}{1095}    \\  
        & Ease        & 458 & \stackedbar{282}{176}   & 457  & \stackedbar{281}{176}   & 450  & \stackedbar{274}{176}    \\      
        & Relaxed     & 3742 & \stackedbar{2532}{1210}  & 3654  & \stackedbar{2476}{1178}  & 2992 & \stackedbar{1952}{1040}   \\    
        & Satisfied   & 7791 & \stackedbar{5521}{2270}  & 8559  & \stackedbar{6110}{2449}  & 8027 & \stackedbar{5532}{2495}   \\    
        & Serene      & 14 & \stackedbar{5}{9}     & 13   & \stackedbar{5}{8}    & 13 & \stackedbar{5}{8}    \\
        \hline
        \multirow{7}{*}[-2.4ex]{\rotatebox[origin=c]{90}{\scriptsize Low-Valance High-Arousal}} & Afraid      & 4 & \stackedbar{0}{4}  & 4  & \stackedbar{0}{4}    & 4 & \stackedbar{0}{4}   \\
        & Alarmed     & 524 & \stackedbar{194}{330}  & 500  & \stackedbar{189}{311}   & 518 & \stackedbar{191}{327}    \\
        & Angry       & 375 & \stackedbar{125}{250}  & 367  & \stackedbar{122}{245}    & 368 & \stackedbar{122}{246}    \\
        & Annoyed     & 674 & \stackedbar{318}{356}  & 666  & \stackedbar{315}{351}   & 626  & \stackedbar{286}{340}   \\
        & Distressed  & 74 & \stackedbar{24}{50}    & 270  & \stackedbar{76}{194}   & 73  & \stackedbar{23}{50}    \\
        & Frustrated  & 70 & \stackedbar{21}{49}    & 70  & \stackedbar{21}{49}    & 159 & \stackedbar{75}{84}    \\
        & Tense       & 715 & \stackedbar{346}{369}   & 706  & \stackedbar{341}{365}   & 711 & \stackedbar{343}{368}    \\
        \hline  
        \multirow{8}{*}[-4.2ex]{\rotatebox[origin=c]{90}{\scriptsize Low-Valance Low-Arousal}} & Bored       & 107 & \stackedbar{56}{51}    & 106 & \stackedbar{55}{51}    & 106 & \stackedbar{55}{51}    \\
        & Depressed   & 742 & \stackedbar{295}{447}   & 656  & \stackedbar{268}{388}   & 710  & \stackedbar{284}{426}    \\
        & Droopy      & 24 & \stackedbar{4}{20}     & 23   & \stackedbar{4}{19}    & 23 & \stackedbar{4}{19}    \\
        & Gloomy      & 149 & \stackedbar{57}{92}    & 134  & \stackedbar{49}{85}    & 144 & \stackedbar{54}{90}    \\
        & Miserable   & 256 & \stackedbar{87}{169}   & 241  & \stackedbar{84}{157}   & 248 & \stackedbar{84}{164}    \\
        & Sad         & 1948 & \stackedbar{813}{1135}  & 1852  & \stackedbar{772}{1080}  & 1923 & \stackedbar{801}{1122}    \\
        & Sleepy      & 1 & \stackedbar{0}{1}     & 1   & \stackedbar{0}{1}    & 1 & \stackedbar{0}{1}    \\
        & Tired       & 38 & \stackedbar{17}{21}    & 36  & \stackedbar{17}{19}    & 38 & \stackedbar{17}{21}    \\

        \bottomrule
    \end{tabular}
    }
\end{table*}

Next, a set of extensive experiments was conducted to optimize the MLP model. 
The models were tested with two to four layers, with \textit{dropout rates} ranging from 0.1 to 0.3 and \textit{learning rates} of 0.001 to 0.0005. In total, 32 models were trained during this process. 
The embeddings used as input was Word2Vec + fastText + SinBERT + SBERT 
all of which were preprocessed and converted into fixed-size arrays (768 dimensions). The top five best performing parameter combinations are summarized in Table~\ref{tab:mlp_opsperformance}.
%
%
The best-performing model was a three-layer MLP with a configuration of 256, 128, and 64 neurons, achieving a \textit{ROC-AUC} score of 0.887. This architecture demonstrated superior performance in terms of both classification accuracy and model balance, outperforming other hyperparameter combinations.

\subsection{Sinhala YouTube Comment Classification}

Finally, using the above best performing model as our pre-trained model, a zero-shot positive-negative classification of the YouTube Sinhala Comment data set for Sinhala songs was conducted, which is shown in the Table~\ref{tab:mlp_classification}. According to the table, all comments classified by the annotators as \textit{Afraid} and \textit{Sleepy} were categorized as Negative by the MLP model. In contrast, comments related to other emotions were a mix of Negative and Positive classifications. 

\section{Conclusion}

This research examined emotion recognition in Sinhala song comments, creating a high-quality annotated data set of 63,471 comments annotated using Russell's V-A model. The comprehensive annotation process used in the study, which has provided an 84.96\% agreement, ensures reliability of the data set generated from the study. 
Subsequent analysis discovered distinct emotional profiles for different Sinhala songs on YouTube, highlighting the importance of comment-based emotion mapping. The study addressed the complexities and challenges of comparing comment-based and song-based emotions, while trying to reduce the biases in user-generated content.
Experiments conducted by~\citet{ranathunga2021sentiment} on Sinhala News Comment classification were used as an expanded baseline for feature selection and model selection whereby the MLP model outperformed other models. Further hyperparameter fine tuning on the MLP model was carried out before utilizing the same model for sentiment classification in Sinhala YouTube comments.   
While this study achieves reasonable performance, future work could explore advanced deep learning architectures. For example, developing the current binary classifier into a multi-class model to better suit the Sinhala YouTube comment dataset. This study's annotated dataset and insights contribute significantly to Sinhala NLP and provide a solid foundation for future music emotion recognition research in low-resource languages. 
%


%
%
%
%
\bibliographystyle{splncs04nat}
\bibliography{References}

\appendix

\section{Hyperparameters for Classification Models}
\label{app:Hype}

For LR, the solver \textit{"saga"} was chosen 
with \textit{L1} and \textit{L2} regularization. The \textit{inverse regularization strength (C)} was set to 3
while the maximum number of iterations was increased to 5000.
Similarly, SVM was trained with a \textit{linear kernel}.
Furthermore, \textit{probability=True} was enabled to allow probability-based predictions.
Random Forest was configured with 300 trees 
and a \textit{maximum depth} of 30
Meanwhile, XGBoost was optimized with 400 trees 
and a maximum depth of 10.
A \textit{learning rate} of 0.1.
Additionally, \textit{eval-metric='mlogloss'} was used to monitor the model's performance. 
The MLP model utilized a fully connected network with four hidden layers (512, 256, 128, 64): \textit{ReLU activation}:llayer normalization.
The LSTM model included three stacked LSTM layers (256, 128, 64):layer normalization:dropout regularization.
Similarly, the CNN model comprised three \textit{CONV layers} with 256, 128, and 64 filters, followed by a \textit{global max pooling layer} and two dense layers. 
%

\end{document}